\documentclass[journal]{IEEEtran}
\ifCLASSINFOpdf
  \usepackage[pdftex]{graphicx}
\else
\fi
\usepackage{amsmath}
\usepackage{algorithmic}

\usepackage{array}
\usepackage[hoptionsi]{overpic}
\DeclareMathOperator*{\argmin}{arg\,min}
\usepackage{amssymb}  
\usepackage{pict2e}

\usepackage{units}
\usepackage{xcolor}
\usepackage{xspace}
\usepackage{multirow}
\usepackage{booktabs}
\usepackage{subcaption}

\usepackage{soul}

\newcommand{\highlight}[1]{\textit{#1}}

\begin{document}
%
\title{Dynamic collision avoidance for multiple robotic manipulators based on a non-cooperative multi-agent game}
%
%
%

\author{Nigora~Gafur,
        Gajanan~Kanagalingam
        and~Martin~Ruskowski
\thanks{The authors are with the Chair of Machine Tools and Control Systems, Department of Mechanical and Process Engineering, Technische Universität Kaiserslautern and German Research Center for Artificial Intelligence (DFKI), Kaiserslautern D-67663, Germany (e-mail: nigora.gafur@mv.uni-kl.de)}}
\maketitle

\begin{abstract}
A flexible operation of multiple robotic manipulators in a shared workspace requires an online trajectory planning with static and dynamic collision avoidance. In this work, we propose a real-time capable motion control algorithm, based on non-linear model predictive control, which accounts for static and dynamic collision avoidance. The proposed algorithm is formulated as a non-cooperative game, where each robot is considered as an agent. Each agent optimizes its own motion and accounts for the predicted movement of surrounding agents. We propose a novel approach for collision avoidance between multiple robotic manipulators. Additionally, we account for deadlocks that might occur in a setup of multiple robotic manipulators. We validate our algorithm on multiple pick and place scenarios and different numbers of robots operating in a common  workspace in the simulation environment Gazebo. The robots are controlled using the Robot Operating System (ROS). We demonstrate, that our approach is real-time capable and, due to the distributed nature of the approach, easily scales up to four robotic manipulators with six degrees of freedom operating in a shared workspace.
\end{abstract}


%
\IEEEpeerreviewmaketitle
\section{Introduction}
\begin{figure}
    \centering
    \includegraphics[width=0.49\textwidth]{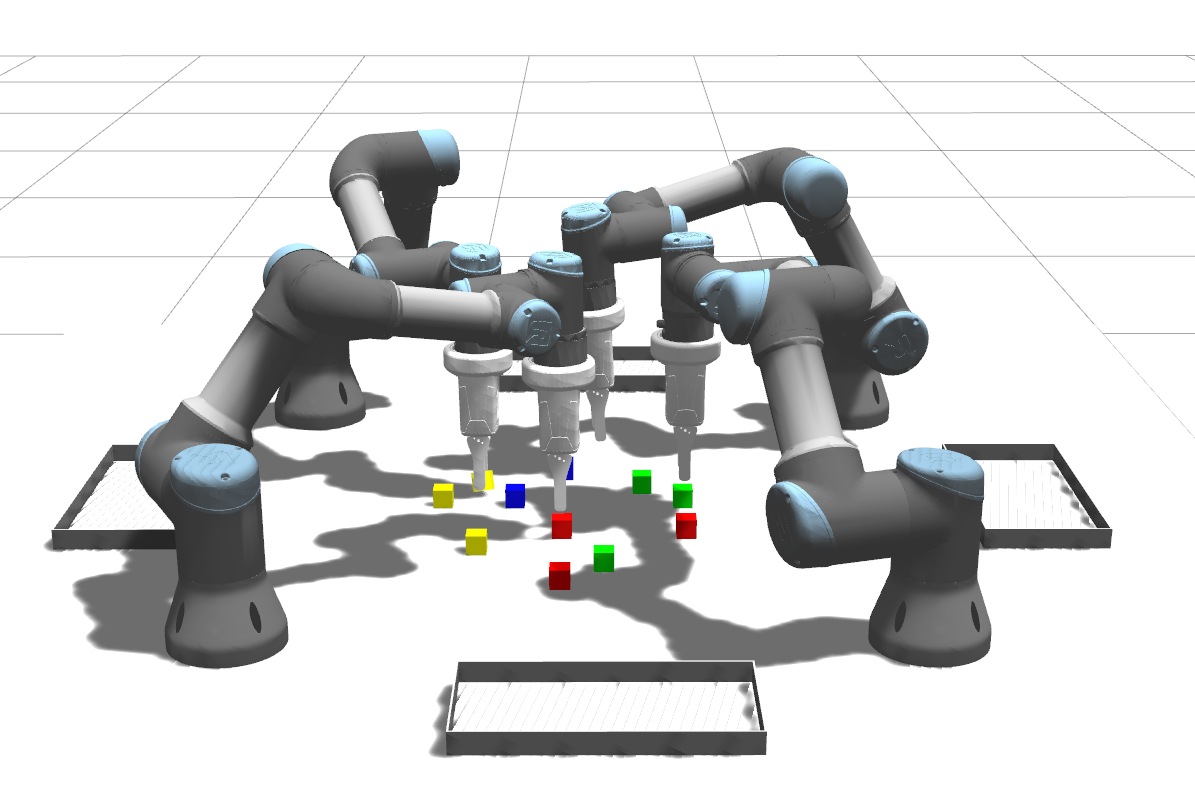}
    \caption{Setup for a pick and place scenario with four collaborative UR$3$ manipulators.}
    \label{fig:4Robots}
\end{figure}

Modern industrial processes are increasingly dominated by shorter innovation and product life cycles, reflecting a growing demand for customized products \cite{jesko2020}. Consequently, factory systems must become more flexible and adaptable \cite{ruskowski2020,wrede2016vertical}. Robotic manipulators are capable of providing such flexibility due to their complex kinematic chain. Areas of application are, for instance, assembly, disassembly or packaging lines. Operating in a shared workspace, several robotic manipulators can further increase efficiency, minimize the working area and make collaboration possible. Figure~\ref{fig:4Robots} constitutes an example of four robotic manipulators sharing the same workspace and performing a pick and place task.\\

Traditionally, the collision free trajectories of all involved robotic manipulators in industrial applications are planned for a specific task involving an unchanging environment. 
As robotic manipulators are generally deployed for repetitive tasks, it suffices to plan collision free trajectories only once. In case certain parts of the production process are changed, a re-planning of collision free trajectories and re-programming of all involved manipulators is necessary. For that reason, it is imperative to develop efficient, scalable and real-time capable motion control strategies which enable a safe and flexible operation of multiple manipulators in changing environments. Such strategies would enable, for example, an on-demand task assignment in a multi-robot setting. Furthermore, modular approaches are conceivable, where each robot may be considered as an independent module. The ability to couple and rearrange such modules in a flexible way would be highly desirable from the point of view of modern production processes.

\section{Contribution and outline} \label{sec:problem_description}
This article is concerned with developing an online motion control algorithm which enables several manipulators to operate simultaneously in a common workspace. We formulate the problem of online motion control for each manipulator as an optimization problem in the joint space, which incorporates static and dynamic collision avoidance constraints. To this end, we derive a novel approach for collision avoidance between multiple robots which enables a safe robot-robot interaction. Our approach is based on MPC to account for disturbances and uncertainties during motion control. Moreover, we use the predictive nature of MPC to exchange information between the robots and thus to account for collisions a priori. We take special care to ensure that our approach is real-time capable. To this end, we use the concept of distributed model predictive control (DMPC) in the joint space, based on a non-cooperative game, where each robotic manipulator is considered as an agent and shares the predicted trajectory with its neighbours. Collision avoidance between robots is not sufficient to overcome the problem of deadlocks. To this end, we introduce a concept how deadlocks among two or more robotic manipulators can be detected and, subsequently, resolved. 

To demonstrate the efficiency of our approach, we consider a setup of multiple $6$-degrees of freedom robotic manipulators in the simulation environment \textit{Gazebo} \cite{koenig2004design}, controlled by ROS. The robotic manipulators are closely placed to each other and operate in a common workspace. We assign each robot several pick and place tasks. The robots and objects are placed in such a way that collisions and deadlocks between the robots are imminent. 
We propose a modular approach, where each robot is considered as an independent module with the ability to cooperate with several other robots by coupling several modules with each other. This approach has the advantage of realizing different setups of multi-robot systems depending on how many robots and what constellation of robots are required to fulfill a task. Further, we compare our approach with sampling-based and optimization-based planners to show the efficacy of our approach. Last but not least, we compare computation times for different setups and draw conclusions about scalability of our approach. 

The remainder of this article is organized as follows. In Section \ref{sec:related_work} we elaborate on existing trajectory generation methods and multi-robot planners. In Section \ref{sec:dynamic_model} the dynamic model of a robotic manipulator is introduced, followed by a formulation of the DMPC problem in Section \ref{sec:DMPC}. A novel approach for collision avoidance is explained in detail in Section \ref{sec:collision_avoidance}. Further, we introduce a novel approach for deadlock detection and resolution in Section \ref{sec:deadlock}. Validation of our algorithm and simulation results are shown in Section \ref{sec:results} followed by a conclusion in Section \ref{sec:conclusion}.

\section{Related Work} \label{sec:related_work}
Motion planning is still an on-going and challenging research area in robotics. In industrial applications, trajectory generation of manipulators is usually required, in addition to its feasibility, to minimize certain criteria, such as the distance travelled or traveling time, and maximizing others, such as energy efficiency or performance. In addition, considering dynamically changing environment is necessary to allow for a flexible operation of a manipulator. In multi-robot systems, each robot has to find a feasible path in a complex and constantly changing environment while sharing its workspace with other robots. In general, the applied methods for trajectory generation in robotic applications can be divided into two main categories: sampling-based and control-based methods. 

Sampling-based methods include the well-known and widely used algorithms based on either rapidly exploring random trees (RRT's) \cite{lavalleRRT} or probabilistic roadmaps (PRM's) \cite{Kavraki1996}. The RRT method is realized as a multi-query planner, whereas the PRM method is a single-query planner \cite{kavraki2016motion}. The sampling-based planners are suitable for high-dimensional configuration spaces and thus for multi-robot systems, which is the main advantage of the methods. Several sampling-based approaches exist for multi-robot motion planning, such as discrete RRT (dRRT) \cite{solovey2016finding} and subdimensional expansion \cite{wagner2015subdimensional}. Recently, an  asymptotically-optimal extension of dRRT was introduced denoted as dRRT*, that was successfully applied for $4$ robotic arms, each with $7$ degrees of freedom sharing a common workspace \cite{shome2020drrt}.  However, the sampling-based methods are mainly applied for static environments, as the trajectories are first planned for a specific task and thereafter executed. The methods are therefore mainly used for offline trajectory planning. Further limitation of the sampling-based method includes difficulties in planning trajectories for narrow passages that often lead to jerky and unnecessary motions \cite{geraerts2007creating}. The Open Motion Planning Library (OMPL) \cite{OMPL} includes a large variety of sampling-based planners which are also integrated in the Robot Operating System (ROS) \cite{ros} framework. 

Control-based planners require a more tailor-made approach depending on the type of robot. This category includes artificial potential fields \cite{khatib1986real} and optimization-based approaches \cite{kalakrishnan2011stomp}, \cite{ratliff2009chomp}, \cite{schulman2013finding}. Both methods search for a feasible path towards the goal based on local information from the environment. The artificial potential field method uses a potential function that induces repulsive forces against obstacles and attractive forces towards the goal. Wang et al. \cite{Wang2016} applied this method for a space manipulator with multiple obstacles occupying the same workspace. Obstacles are only considered if the manipulator undercuts a predefined minimal distance to the individual objects. Bosscher et al. \cite{Bosscher2011} applies velocity damping for a cooperative motion planning of two robotic manipulators, where a trajectory is planned for each robot in advance and collisions are considered only during the execution of the trajectory. The main drawback of the potential field method is its limitation to a low-dimensional configuration space. 

Optimization-based methods are usually formulated as constrained optimization problems. The feasibility of the trajectory is ensured by incorporating a kinematic and a dynamic model of the corresponding robot in the constraints of the optimization problem. Additionally, static as well as dynamic obstacles may be considered by adding additional constraints to the optimization problem. A large number of constraints can result in high computational burden. Therefore, an efficient incorporation of constraints is required, especially for multi-robot systems in a dynamically changing environment. The concept of model predictive control (MPC) \cite{Mayne2000} in a receding horizon formulation is suitable for solving the trajectory generation problem for a dynamically changing environment by formulating an optimization problem that is solved over a prediction horizon. The main advantage of the MPC framework is its predictive nature that gives an insight on the future trajectory. Exchange of information with other robots enables to account for potential collisions a priori. Further, the closed-loop control accounts for model uncertainties and disturbances.

Trajectory generation with MPC for a single robotic manipulator without collision avoidance was carried out by Lam et al. \cite{Lam2013}, Arkadani et al. \cite{Ardakani2015} and Belda et al. \cite{Belada2017}. There are two possible approaches integrating collision avoidance into trajectory generation with MPC. In the first approach, the MPC algorithm itself is extended by solving the optimization problem not over the whole state space of the considered system, but only over a subset of the state space. This subset excludes all states where a collision might occur and needs to be determined a priori. This method was applied by Liu et al.~\cite{Liu2018} and Schoels et al.~\cite{Schoels2020} for trajectory generation of a mobile robot, where Schoels et al.~\cite{Schoels2020} approximated the collision free subset by circles and Liu et al.~\cite{Liu2018} used polyhedra at the current state. Rösmann et al. \cite{rosmann2015planning} uses a global planner to optimize trajectories of multiple mobile robots. An extension of these approaches to manipulators is not known to the authors.


The second approach to integrate collision avoidance into trajectory generation with MPC is to introduce further constraints to the optimization problem. There exist several approaches to formulate these constraints. One approach is to restrict the distance of all collision-prone object pairs, where the corresponding objects are approximated by convex bodies. Thus, for a kinematic model of a manipulator this results in a connected chain of convex bodies \cite{Bosscher2011}, \cite{Cascio2009}, \cite{Kraemer2020}, where each pair of collision-prone bodies introduces an additional constraint into the optimization problem, e.g., in case of multiple manipulators or a manipulator and a human.

The computation of distances between two convex bodies is done by algorithms with nested logical conditions \cite{Lumelsky1985}, \cite{Gilbert1988}, \cite{Rimon1997}. However, the derivatives of the constraints are not smooth, which poses additional challenges to solving the underlying optimization problem. Krämer et al. \cite{Kraemer2020} extends the collision avoidance approach from Lumelsky \cite{Lumelsky1985} and proposes an online motion control for one robotic manipulator in collaboration with a human. The computation times prove the efficacy of the approach, where the optimization problem is solved with a self developed hypergraph \cite{rosmann2018exploiting} to mitigate the problem of nested logical conditions. 

As an alternative to restricting the distance, virtual hyperplanes can be used to separate two collision-prone bodies. By approximating the considered objects by polyhedra and applying Farkas' lemma, collisions of the considered objects can be avoided. An implementation with multidimensional polyhedra was proposed by Gerdts et al. for a robotic manipulator \cite{Gerdts2012}. In the work of Zhang et al.~\cite{Zhang2020} this approach is extended so that, in addition to collision avoidance, a minimum distance between two bodies can be guaranteed. The former approach comes with the disadvantage, that for every pair of collision-prone objects, several constraints have to be added to the underlying optimization problem. Six new optimization variables have to be introduced into the optimization problem for each considered object pair.  The number of additional constraints depends linearly on the number of polyhedron faces, which is computationally intractable for multi-robot systems. 

The framework of MPC can be realized in a centralized or distributed fashion. The drawback of the centralized MPC is the limited scalability and high computational complexity \cite{christofides2013distributed}. A distributed MPC framework in the context of game theory can help to split the computational burden, where each agent optimizes its own objective function \cite{christofides2013distributed}. This concept was already introduced for robot-human collaboration by Flad et al. \cite{flad2017cooperative}. Yanhao et al. \cite{yanhao21} proposes an approach based on a distributed control for a cooperative manipulation of an object. Tika et al. applied centralized MPC \cite{tikacentral} and distributed MPC \cite{tikaDMPC} for a synchronous pick and place scenario for two robotic manipulators. However, the focus lies on a synchronous task accomplishment for two robotic manipulators rather than collision avoidance. Furthermore, deadlocks are not treated in any of the mentioned works. Existing approaches, still, cannot guarantee a collision free trajectory generation in dynamically changing environments, that is real-time capable and scales to more than two manipulators. 

\section{Dynamic model} \label{sec:dynamic_model}
We consider a robotic manipulator with $N$ joints, where each joint is actuated by a servomotor with high transmission ratio. Thus, with a decentralized control scheme it leads to a system dynamics of $N$ double integrators where each joint is independently controlled \cite{bruno2010robotics}. The dynamic model of a manipulator admits the representation
\begin{equation}
\label{eq:linearRobotDynamics}
\begin{bmatrix}
 \dot{\mathbf{q}}(t) \\  \ddot{\mathbf{q}}(t)
\end{bmatrix}= 
\begin{bmatrix}
\mathbf{0} & \mathbf{I} \\
\mathbf{0} & \mathbf{0} 
\end{bmatrix}
\begin{bmatrix}
\mathbf{q}(t) \\ \dot{\mathbf{q}}(t)
\end{bmatrix}+
\begin{bmatrix}
\mathbf{0} \\ \mathbf{I}
\end{bmatrix} \mathbf{u}(t),
\end{equation}
where each joint of a manipulator is independently controlled. ${\mathbf{q}(t)\in\mathbb{R}^N}$ denotes the joint angular position vector, $\mathbf{u}(t) \in \mathbb{R}^N$ is the control input vector, the matrix $\mathbf{I} \in \mathbb{R}^{N \times N}$ denotes the identity matrix and  $\mathbf{0} \in \mathbb{R}^{N \times N}$ represents the zero matrix. 

We derive a discrete-time representation of the linear system with the state vector $\mathbf{x}(t)=[\mathbf{q}(t), \ \dot{\mathbf{q}}(t)]^T \in \mathbb{R}^{2N}$ in the state-space 
\begin{equation}
\label{eq:state_space}
    \mathbf{x}^{k+1} = \mathbf{A}^d\mathbf{x}^k + \mathbf{B}^d\mathbf{u}^k,
\end{equation}
where $\mathbf{A}^d \in \mathbb{R}^{2N \times 2N}$ represents the discrete state matrix and $\mathbf{B}^d \in \mathbb{R}^{2N \times N}$ is the input matrix. The equation \eqref{eq:state_space} is discretized with a sample time $T_\textrm{s}$, where $(\cdot)^k$ represent discrete variables at time $t_k=k\cdot T_s$. The discrete states are denoted in the following as $\mathbf{x}_i^k = \mathbf{x}_i(t_k)$ and discrete control inputs as $\mathbf{u}_i^k = \mathbf{u}_i(t_k)$ for a manipulator $i$.
The linear system in \eqref{eq:linearRobotDynamics} describes the dynamics of a robotic manipulator in the joint space, which will be integrated as a constraint together with static and dynamic collision avoidance constraints into an optimization problem. This will be discussed in more detail in Sections \ref{sec:DMPC} and \ref{sec:collision_avoidance}.

\section{Distributed model predictive control in the context of game theory} \label{sec:DMPC}
We consider a system of $M$ robotic manipulators. Each manipulator ${i=1, \dots, M}$ represents an independent subsystem. The main objective of performing a cooperative task is, for every robotic manipulator, to safely reach the target joint state accounting for static, dynamic and self- collision constraints. 


Centralized MPC considers the overall system dynamics in a single optimization problem with respect to a common objective function $J$, which can be written as
\begin{gather} 
    \mathbf{u}^{* \ 0:N_\textrm{p}-1}_1, \dots, \mathbf{u}^{* \ 0:N_\textrm{p}-1}_M = \argmin_{\mathbf{u}_1, \dots, \mathbf{u}_M} \ J(\mathbf{u}^{0:N_\textrm{p}-1}_1, \dots, \mathbf{u}^{0:N_\textrm{p}-1}_M),
\end{gather}
where $N_\textrm{p}$ denotes the prediction horizon. For brevity we choose the following notation for the control inputs $\mathbf{u}^{0:N_\textrm{p}-1}_i = [\mathbf{u}^0_i, \dots, \mathbf{u}^{N_\textrm{p}-1}_i]$. However, the degrees of freedom of the former approach increase with an increasing number of robots such that the computational cost quickly become inadmissible for real-time applications. In this work, we investigate distributed model predictive control (DMPC),  where each robot is considered as an agent. There is a multitude of different architectures for distributed model predictive control \cite{christofides2013distributed}. From the point of view of game theory and by classification via the cost function, DMPC can be realized as a cooperative or a non-cooperative game. Both games rely upon communication between the agents. \\

In a cooperative game, all agents optimize a \highlight{global} cost function $J$, i.e.\ the agents share a common objective. Cooperative agents negotiate until they agree upon a strategy that brings the best benefit to all of them and is formulated as follows
\begin{equation}
    \begin{gathered} 
        \mathbf{u}^{* \ 0:N_\textrm{p}-1}_1 = \argmin_{\mathbf{u}_1^{0:N_\textrm{p}-1}} \ J(\mathbf{u}^{0:N_\textrm{p}-1}_1, \mathbf{u}^{* \ 0:N_\textrm{p}-1}_2, \dots, \mathbf{u}^{* \ 0:N_\textrm{p}-1}_M),\\
        \vdots \\
        \mathbf{u}^{* \ 0:N_\textrm{p}-1}_M = \argmin_{\mathbf{u}_M^{0:N_\textrm{p}-1}} \ J(\mathbf{u}^{* \ 0:N_\textrm{p}-1}_1, \dots, \mathbf{u}^{* \ 0:N_\textrm{p}-1}_{M-1}, \mathbf{u}^{0:N_\textrm{p}-1}_M).
        \label{eq:coop_DMPC}
    \end{gathered}
\end{equation}

In a non-cooperative setting, the agents pursue their own goal and therefore act egoistically to achieve their own best possible benefit. Each agent optimizes its own, i.e. \highlight{local} cost function $J_i$,  $i = 1, \dots, M$. The non-cooperative game has the following form
\begin{equation}
    \begin{gathered} 
        \mathbf{u}^{* \ 0:N_\textrm{p}-1}_1 = \argmin_{\mathbf{u}_1^{0:N_\textrm{p}-1}} \ J_1(\mathbf{u}^{0:N_\textrm{p}-1}_1, \mathbf{u}^{* \ 0:N_\textrm{p}-1}_2, \dots, \mathbf{u}^{* \ 0:N_\textrm{p}-1}_M),\\
        \vdots \\
        \mathbf{u}^{* \ 0:N_\textrm{p}-1}_M = \argmin_{\mathbf{u}_M^{0:N_\textrm{p}-1}} \ J_M(\mathbf{u}^{* \ 0:N_\textrm{p}-1}_1, \dots, \mathbf{u}^{* \ 0:N_\textrm{p}-1}_{M-1}, \mathbf{u}^{0:N_\textrm{p}-1}_M).
        \label{eq:noncoop_DMPC}
    \end{gathered}
\end{equation}
The non-cooperative game converges towards a Nash equilibrium \cite{christofides2013distributed}, whereas the optimal solution of a cooperative game is Pareto optimal \cite{stewart2010cooperative}. \\

In general, systems considered in a cooperative \eqref{eq:coop_DMPC} and a non-cooperative framework \eqref{eq:noncoop_DMPC} are coupled in control inputs, such that each subsystem can not be optimized independently without knowledge of the optimal strategies of other agents. For robotic manipulators working independently in a shared workspace, as e.g. pick and place tasks, the system dynamics are decoupled in states and  control inputs. A coupling of the robots' system dynamics occurs e.g., if robotic manipulators are physically attached to each other, which we do not consider in this work. \\

\subsection{Formulation of DMPC problem for a non-cooperative multi-agent game}
In this work, the robots are solely coupled in states by the respective collision avoidance constraints. The local cost functions $J_i$ of each agent still remain  decoupled in states and control inputs with regard to other agents. The drawback of cooperative DMPC is that each \highlight{local} controller has to have knowledge of the full system dynamics and several communication iterations are needed until an optimal solution for the whole game is obtained. Therefore, we consider a non-cooperative game in the following, which has the benefit of local subsystems and local cost functions as well as single communication iteration at each time step. \\

Keeping the former in mind, we turn our attention to the formulation of the online trajectory planning problem based on DMPC, formulated as a non-cooperative game. We choose the multiple shooting method for discretizing the optimization problem. The prediction horizon $N_\textrm{p}$ is split equidistantly into $t_k = k \cdot T_\textrm{s}$ time steps with $k=0,\cdots, N_\textrm{p}$. The DMPC formulation for each involved robotic manipulator $i$ takes the following form 
\begin{subequations}\label{eq:DMPC}
\begin{align}
         {\underset{\mathbf{u}_{i}^{0:N_\textrm{p}-1}, \mathbf{x}_{i}^{0:N_\textrm{p}}}{\min}} \quad & J^\textrm{f}_i(\mathbf{x}_i^{N_\textrm{p}}) + \sum_{k=0}^{N_\textrm{p}-1} J^\textrm{c}_i(\mathbf{x}^k_{i},\mathbf{u}^k_{i}) 
        \tag{\ref{eq:DMPC}} \\
    	s.t. \quad &     \mathbf{x}^{k+1}_i = \mathbf{A}^\textrm{d}_i\mathbf{x}^k_i + \mathbf{B}^\textrm{d}_i\mathbf{u}^k_i, \quad k=0,...,N_\textrm{p}-1 , \label{eq:dynamic_k}\\
    	& \mathbf{x}^0_i = \mathbf{x}^\textrm{s}_i \label{eq:xinit_k}, \\
        & \mathbf{x}^k_i \in \bar{\mathbb{X}}_i , \quad k=0,...,N_\textrm{p}-1 \label{eq:x_lim_k},\\
        & \mathbf{u}^k_i \in \bar{\mathbb{U}}_i , \quad k=0,...,N_\textrm{p}-1 \label{eq:u_lim_k},\\
    	& R_i(\mathbf{x}^k_i) \cap \mathcal{O}  = \varnothing  , \quad k=0,...,N_\textrm{p} \label{eq:static_collision_k},\\
        & R_i(\mathbf{x}^k_i) \cap \mathcal{R}_{-i}(\underline{\mathbf{x}}^{* \; k}_{-i})  = \varnothing  , \quad k=0,...,N_\textrm{p} . \label{eq:dynamic_collision_k}
\end{align}
\end{subequations}

The quadratic cost function $J^\mathrm{c}_i : \mathbb{R}^{2N} \times \mathbb{R}^{N} \to \mathbb{R}$,
\begin{equation} \label{cost_func_init}
    \begin{split}
            J^\textrm{c}_i(\mathbf{x}^k_{i},\mathbf{u}^k_{i})  := & (\mathbf{x}^k_{i}-\mathbf{x}_i^\mathrm{f} )^\mathrm{T}\mathbf{Q}^x_i(\mathbf{x}^k_{i}-\mathbf{x}_i^\mathrm{f}) +  \\ 
            & \mathbf{u}^{k \; \mathrm{T}}_{i}\mathbf{R}^\mathrm{u}_i\mathbf{u}^k_{i} +\Delta \mathbf{u}_i^{k \; \mathrm{T}}\mathbf{R}^\mathrm{d}_i\Delta \mathbf{u}_i^{k}
    \end{split}
\end{equation}
penalizes the squared state error, i.e., the deviation of the state $\mathbf{x}^k_{i}$ from the desired state $\mathbf{x}_i^\mathrm{f}= [\mathbf{q}^\mathrm{f \; T}_i, \ \mathbf{0}^{ \; \mathrm{T}}]^\mathrm{T}$, the magnitude of the control input $\mathbf{u}^{k}$ and the control smoothness, i.e., the magnitude of $\Delta \mathbf{u}_i^{k} = \frac{\mathbf{u}_i^{k+1}-\mathbf{u}_i^{k}}{t_{k+1}-t_k}$, with the positive (semi-) definite weighting matrices $\mathbf{Q}^x_i \in \mathbb{R}^{2N \times 2N}$, $\mathbf{R}^\mathrm{u}_i \in \mathbb{R}^{N \times N}$ and $\mathbf{R}^\mathrm{d}_i \in \mathbb{R}^{N \times N}$, respectively. The terminal state cost $J^\mathrm{f}_i : \mathbb{R}^{2N} \to \mathbb{R}$
\begin{equation} \label{cost_func_terminal}
    J^\textrm{f}_i(\mathbf{x}_i^{N_\textrm{p}}) :=  (\mathbf{x}^{N_\textrm{p}}_{i}-\mathbf{x}^\mathrm{f}_i)^\mathrm{T}\mathbf{Q}^\mathrm{f}_i(\mathbf{x}_i^{N_\textrm{p}}-\mathbf{x}^\mathrm{f}_i)
\end{equation}
penalizes the terminal squared state error with the positive (semi-) definite weighting matrix $\mathbf{Q}^\mathrm{f}_i \in \mathbb{R}^{2N \times 2N}$. \\

The dynamics of manipulator $i$ is given by equation \eqref{eq:dynamic_k}, see Section \ref{sec:dynamic_model}, whereas the equation \eqref{eq:xinit_k} sets the measured joint state $\mathbf{x}_i^\mathrm{s}$ of manipulator $i$ as the initial condition of the state vector $\mathbf{x}^k_i $ at time $k=0$. The equations \eqref{eq:x_lim_k} and \eqref{eq:u_lim_k} represent lower and upper bounds on the states and control inputs, i.e.,\ 
\begin{equation}
        \begin{split}
            \bar{\mathbb{X}}_i &:= \{ \mathbf{x}^k_i  \in \mathbb{R}^{2N} \mid  \mathbf{x}_{i,\text{min}} \leq \mathbf{x}^k_i  \leq \mathbf{x}_{i,\text{max}}  \}, \\
            \bar{\mathbb{U}}_i &:= \{ \mathbf{u}^k_i  \in \mathbb{R}^N \mid  \mathbf{u}_{i,\text{min}} \leq \mathbf{u}^k_i  \leq \mathbf{u}_{i,\text{max}} \}.
        \end{split}
        \label{eq:self_col}
\end{equation}
The equation for joint angles \eqref{eq:self_col} also accounts for self-collision constraints through limitation of the joints' angle ranges.\\

We formulate static and dynamic collision avoidance constraints in the task space, transforming joint positions into positions in Cartesian space using the non-linear forward kinematics. The collision avoidance constraints turn the optimization problem \eqref{eq:DMPC} into a non-convex one. We define the set 
\begin{equation}
    \mathcal{R}(\underline{\mathbf{x}}^k) := R_1(\mathbf{x}_1^k) \cup ... \cup R_M(\mathbf{x}_M^k) ,
\end{equation}
where $R_i(\mathbf{x}_i^k)$ denotes the interior set of Cartesian points occupied by manipulator $i$ with state $\mathbf{x}_i^k$. The trajectory vector $\underline{\mathbf{x}}^k = [\mathbf{x}_1^k, \dots, \mathbf{x}_M^k]$ collects the trajectories of all involved robots. To account for static objects in the task space, we define the set $\mathcal{O}$ containing all interior points of all static obstacles. Indeed, constraint \eqref{eq:static_collision_k} enforces that the intersection of $R_i(\mathbf{x}_i^k)$ and the obstacles $\mathcal{O}$ for state $\mathbf{x}_i^k$ is empty. To consider dynamic collision avoidance constraints, i.e., the prevention of inter-robot collisions, the short-hand notation 
\begin{equation}
    \mathcal{R}_{-i}(\underline{\mathbf{x}}_{\; -i}^k) := \bigcup_{j=1,j\neq i}^M   R_j(\mathbf{x}_j^k)
\end{equation}
is introduced. Consequently, constraint \eqref{eq:dynamic_collision_k} prevents the inter-robot collision of robot $i$ with all other robots. The efficient implementation of constraints \eqref{eq:static_collision_k} and \eqref{eq:dynamic_collision_k} is the topic of the following section. In the following, we focus on solving the DMPC before turning to the efficient implementation of the collision avoidance constraints. \\

Note, that constraint \eqref{eq:dynamic_collision_k} establishes the coupling between the manipulators. Constraint \eqref{eq:dynamic_collision_k}
implies that the optimal trajectories of all other robots, collected in $\underline{\mathbf{x}}^{* \; k}_{\; -i}$, is known a priori in order to solve the optimization problem \eqref{eq:DMPC} for manipulator $i$. To obtain 
\begin{equation}
    \underline{{\mathbf{x}}}^{* \; 0:N_\textrm{p}}_{-i} = [{\mathbf{x}}^{* \; 0:N_\textrm{p}}_1, \dots, {\mathbf{x}}^{* \; 0:N_\textrm{p}}_{i-1}, {\mathbf{x}}^{* \; 0:N_\textrm{p}}_{i+1}, \dots, {\mathbf{x}}^{* \; 0:N_\textrm{p}}_M],
\end{equation}
for collision constraint \eqref{eq:dynamic_collision_k} we use an extrapolation approach \cite{christofides2013distributed}. Suppose 
\begin{equation}
    \underline{\hat{\mathbf{x}}}^{* \; 0:N_\textrm{p}} = [\hat{\mathbf{x}}^{* \; 0:N_\textrm{p}}_1, \dots, \hat{\mathbf{x}}^{* \; 0:N_\textrm{p}}_M]
\end{equation}
denotes the manipulators' optimal trajectories from the last converged DMPC-step. We obtain $\underline{\mathbf{x}}^{* \; 0:N_\textrm{p}}$ (and thus also $\underline{{\mathbf{x}}}^{* \; 0:N_\textrm{p}}_{-i}$) by shifting $\underline{\hat{\mathbf{x}}}^{* \; 0:N_\textrm{p}}$ by one time step and extrapolating the last state. In other words, for every manipulator $i=1,\dots,M$, we compute
 \begin{equation}
     \mathbf{x}^{*\;0:N_\textrm{p}}_i = [\hat{\mathbf{x}}^{* \; 1:N_\textrm{p}}_i, \quad \mathbf{x}^{*\;N_\textrm{p}}_i].
 \end{equation}
where the last predicted optimal state $\mathbf{x}_i^{*N_\textrm{p}}$ is obtained by the extrapolation of $\hat{\mathbf{x}}^{*\;N_\textrm{p}}_i$ using the discrete system dynamics, i.e., 
\begin{equation} \label{eq:extrapolation}
    \mathbf{x}^{*N_\textrm{p}}_i = \mathbf{A}^\textrm{d}_i\hat{\mathbf{x}}^{*\;N_\textrm{p}}_i + \mathbf{B}^\textrm{d}_i\mathbf{u}^{*\;N_\textrm{p}-1}_i. 
\end{equation}
Note, that the equation \eqref{eq:extrapolation} can be obtained by setting $\mathbf{u}_i^{*\;N_\textrm{p}-1}=\hat{\mathbf{u}}_i^{*\;N_\textrm{p}-1}$, i.e. the two last optimal inputs in the sequence $\mathbf{u}_i^{*\;0:N_\textrm{p}-1}$ are assumed equal.

To sum up, the model predictive controller of each robot receives its current joint state and the (extrapolated) predicted joint states of neighboured robots $\underline{\mathbf{x}}^{* \; 0:N_\textrm{p}}_{\; -i}$ 
 to account for collisions in the future and choose a proper control strategy to avoid them. 
\subsection{Control Structure} \label{control_structure}
\begin{figure}[h!]
    \centering
    \includegraphics[width=0.49\textwidth]{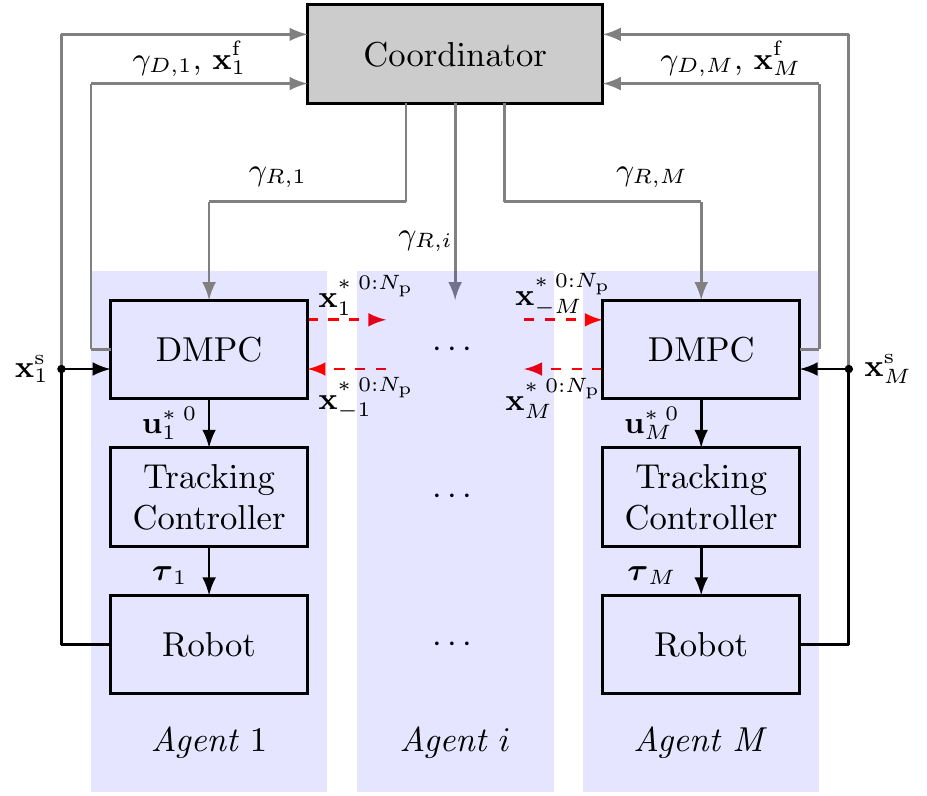}
    \caption{Control structure of collision free online motion control for multiple robotic manipulators.}
    \label{fig:schamtics_system}
\end{figure}

We propose the following control structure of our approach, illustrated in Figure \ref{fig:schamtics_system}. In general, collision avoidance alone is not sufficient to prevent deadlock. A deadlock occurs, when robots prevent each other from reaching their respective target state corresponding to a Nash equilibrium of the corresponding optimization problems. Therefore, a supervisory instance is required to coordinate the robots in order to resolve deadlocks.
The supervisory role is taken over by a coordinator to resolve deadlocks once they are locally detected by  manipulators.  Therefore, communication between the robots and the coordinator is necessary, which is indicated by gray lines in Figure \ref{fig:schamtics_system}.
The coordinator receives a deadlock status denoted as $\gamma_{R,i}$ from each agent whether it is currently in a deadlock. In addition, the coordinator receives the current $\mathbf{x}_i^\textrm{s}$ and the target poses $\mathbf{x}_i^\textrm{f}$. This information is necessary to reliably detect and resolve deadlocks. If manipulators report a deadlock, the coordinator resolves it by sending an activation or deactivation status to each agent denoted as $\gamma_{D,i}$. The DMPCs of the $M$ agents solve the problem  in parallel by accounting for predicted state sequences $\mathbf{x}^{*\;0:N_\textrm{p}}_i$ for $i=1,\dots,M$ of the last converged DMPC step of the neighboured robots.  The  optimal control inputs $\mathbf{u}^{*\;0}_i=\text{const.}$, $i=1,\dots,M$ for $[t_0,\ t_1)$ are sent to the robots' underlying tracking controllers. The robots' controllers generate joint actuator torques $\boldsymbol{\tau}_i(t)$ which are applied to each robots' joints. Subsequently, the current state of a robot $\mathbf{x}^\textrm{s}_i$, $i=1,\dots,M$ is measured and sent to the DMPC. At the same time, the obtained optimal state sequence $\mathbf{x}_i^{* \;0:N_\textrm{p}}$, $i=1,\dots,M$ is communicated to the neighboured robots, indicated by red dashed lines in Figure \ref{fig:schamtics_system}.

\section{Collision avoidance method for multiple robotic manipulators} \label{sec:collision_avoidance}

In the previous section, we formulated the motion control problem of $M$ robotic manipulators as $M$ coupled DMPCs, based on a non-cooperative game. This section is dedicated to the efficient implementation of the static and dynamic collision constraints \eqref{eq:static_collision_k} and \eqref{eq:dynamic_collision_k}.\\

One of the most applied algorithms among collision avoidance methods in the literature is the Lumelsky algorithm \cite{Lumelsky1985}. The method approximates the robot links with line segments and introduces an algorithm to compute the minimum distance between them. The approximation of robot links as line segments is also known as line-swept sphere \cite{Bosscher2011}, \cite{Kraemer2020}, \cite{Cascio2009}. The drawback of the algorithm are the nested logical conditions, which are not smooth and pose challenges to solving the OCP. Therefore, we introduce a novel approach for collision avoidance by approximating a robot's geometry by line segments and ellipsoids and derive an efficient and smooth formulation, that enables the robots to safely avoid collisions. 

\subsection{Ellipsoid - line segment approach} \label{ELS}
\begin{figure}
    \centering
    \includegraphics[width=0.49\textwidth]{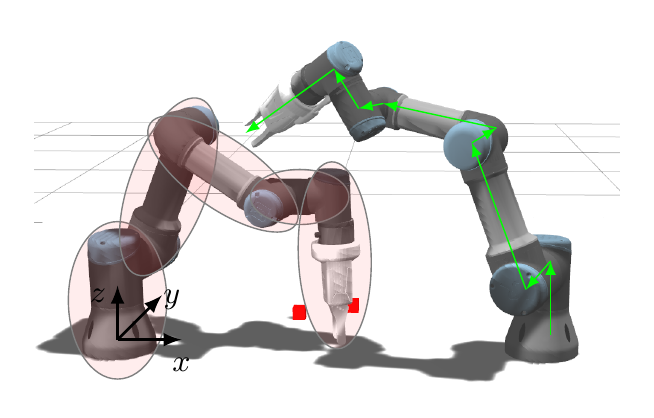}
    \caption{Illustrative approximation of robots' geometry with ellipsoids and line segments from the perspective of the robot $R_i(\cdot)$ on the right side.}
    \label{fig:wireframe_2Robots}
\end{figure}

In order to overcome the problem with nested logical conditions, we do not use a distance function to compute the distance between two links. Instead we ensure at each optimization step that there is no intersection between line segments and ellipsoids formulated as hard constraints in \eqref{eq:dynamic_collision_k}. In the following, we proceed from the perspective of a manipulator $i$ with a set of interior points in the task space denoted by $R_i(\mathbf{x}^k_i)$. The sets of interior points of the other robotic manipulators is designated by $\mathcal{R}_{-i}(\underline{\mathbf{x}}^{* \; k}_{\;-i})$. We approximate the links of a manipulator $i$, for which the optimization problem is solved, by line segments and the links of the remaining manipulators by ellipsoids, we abbreviate it as ELS method. See Figure \ref{fig:wireframe_2Robots} for an illustration, where the robot on the left side is approximated by $5$ ellipsoids while the robot on the right is approximated by $8$ line segments. By choosing proper dimensions of the ellipsoids with suitable safety margin and thus ensuring that the lines and ellipsoids do not intersect, we assure that
\begin{equation}\label{eq:collision_dynamc2}
    R_i(\mathbf{x}^k_i) \cap \mathcal{R}_{-i}(\underline{\mathbf{x}}^{* \; k}_{-i})  = \varnothing  , \quad k=0,...,N_\textrm{p}
\end{equation}
holds. Please note, that all pairs of ellipsoids and line segments of all involved robotic manipulators must be taken into account for every time step $k=0,...,N_\textrm{p}$. \\

In the following we consider collision avoidance between a robot $i$ and a robot $j$, where robot $i$ is modeled with line segments and robot $j$ with ellipsoids. A line segment $\mathbf{s}_m$ of a link $m$ is described by the equation
\begin{equation}
    \mathbf{s}_m(\mathbf{x}^k_i) := \mathbf{b}_m(\mathbf{x}^k_i) + \alpha_m \mathbf{r}_m(\mathbf{x}^k_i), \quad \alpha_m \in [0, 1], \label{eq:line}
\end{equation}
where vector $\mathbf{b}_m(\mathbf{x}^k_i) \in \mathbb{R}^3$ is the position vector of the basis of the considered link and vector $\mathbf{r}_m(\mathbf{x}^k_i)\in \mathbb{R}^3$ designates the direction from $\mathbf{b}_m(\mathbf{x}^k_i)$ to $\mathbf{b}_{m+1}(\mathbf{x}^k_i)$ of the subsequent link. The parameter $\alpha_m$ restricts the line segment to the length of the considered link. \\

For a given state $\mathbf{x}^k_j \in \mathbb{R}^{2N}$, the ellipsoid $\{\mathbf{e} \in \mathbb{R}^{3} \mid H_n(\mathbf{e},\mathbf{x}^k_j) = 1\}$ of a link $n$ is parameterized by the following quadratic equation
\begin{equation}
    H_n(\mathbf{e},\mathbf{x}^k_j) := (\mathbf{e}-\mathbf{e}_{0,n}(\mathbf{x}^k_{j}))^\mathrm{T} \mathbf{R}_n(\mathbf{x}^k_{j}) \mathbf{E}_n \mathbf{R}_n^\mathrm{T}(\mathbf{x}^k_{j}) (\mathbf{e}-\mathbf{e}_{0,n}(\mathbf{x}^k_{j})), \label{eq:ellipsoid}
\end{equation}
where $\mathbf{e} \in \mathbb{R}^3$ names a point on the ellipsoid. The centre point of the ellipsoid is denoted 
\begin{equation}
   \mathbf{e}_{0,n}(\mathbf{x}^k_{j}) = \frac{1}{2}(\mathbf{b}_{n+1}(\mathbf{x}^k_{j}) + \mathbf{b}_{n}(\mathbf{x}^k_{j})),
\end{equation}
the rotation matrix $\mathbf{R}_n(\mathbf{x}^k_{j}) \in \textrm{SO}(3)$ describes the rotation of link $n$ relative to the inertial frame and the diagonal matrix
\begin{equation}
    \mathbf{E}_n = \mathrm{diag}\left(\frac{1}{l_1^2}, \frac{1}{l_2^2}, \frac{1}{l_3^2}\right) \in \mathbb{R}^{3 \times 3}
\end{equation}
contains the squared inverse principal semi-axes $l_1, l_2, l_3 \in \mathbb{R}_{>0}$. To ensure that \eqref{eq:collision_dynamc2} holds, the width of an ellipsoid should be at least twice as large as the width of a robot link and an ellipsoid should also occupy the two joints connecting the link. \\

\begin{figure}[h!]
    \centering
    \includegraphics[width=0.4\textwidth]{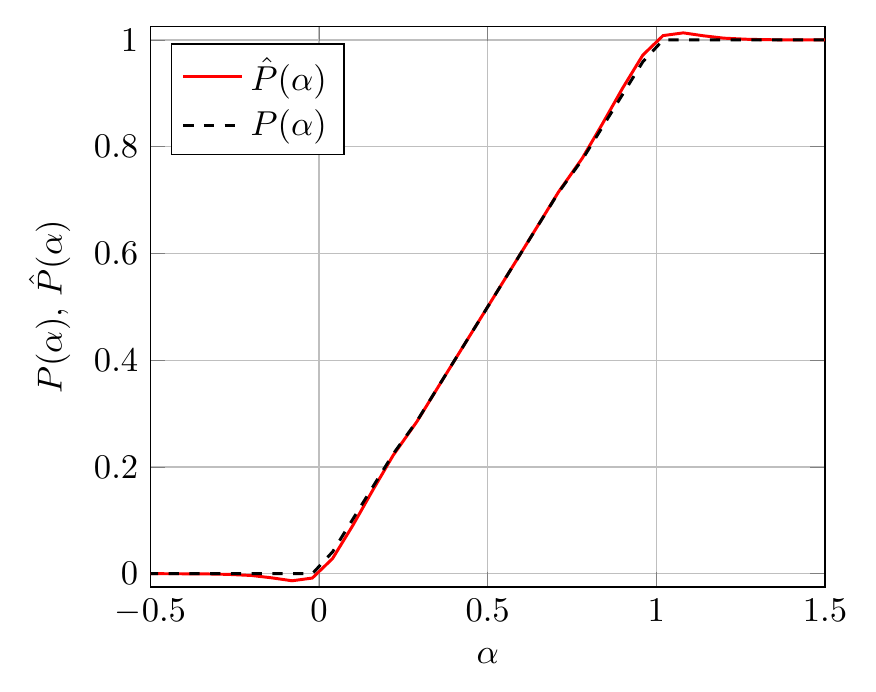}
    \caption{A comparison between the projection operator $P(\alpha)$ and the approximated function $\hat{P}(\alpha)$ with parameter $c=20$.}
    \label{fig:alpha}
\end{figure}

In order to ensure, that line segment $m$ and ellipsoid $n$ do not intersect, the condition
\begin{equation}
    1 - H_n(\mathbf{s}_m(\mathbf{x}^k_i),\mathbf{x}^k_j) \leq 0, \quad \forall \alpha_m \in [0, 1] \label{eq:collision_line_ellipsoid}
\end{equation}
has to hold. Alternatively, the former can be reformulated into an optimization problem, i.e., solving
\begin{subequations}\label{eq:alpha_opt}
    \begin{align}
         \underset{\alpha_m}{\min} \quad & H_n(\mathbf{b}_m + \alpha_m \mathbf{r}_m)
        \tag{\ref{eq:alpha_opt}} \\
    	s.t. \quad & 0 \leq \alpha_m \leq 1 \label{eq:alpha_constr}
    \end{align}
\end{subequations}
for $\alpha_m^*$ where $H_n(\mathbf{b}_m + \alpha^*_m \mathbf{r}_m) \geq 1$ holds. Please note, we dropped explicit reference to $\mathbf{x}^k_i$ and $\mathbf{x}^k_j$ for sake of readability. Problem \eqref{eq:alpha_opt} is solved in the following way. First, the solution $\hat{\alpha}_m \in [-\infty, \infty]$ of the unconstrained optimization problem computes to 
\begin{equation}
    \hat{\alpha}_m = -\frac{(\mathbf{b}_m - \mathbf{e}_{0,n})^\mathrm{T}\mathbf{R}_n\mathbf{E}_n\mathbf{R}_n^\mathrm{T}\mathbf{r}_m}{\mathbf{r}_m^\mathrm{T}\mathbf{R}_n\mathbf{E}_n\mathbf{R}_n^\mathrm{T}\mathbf{r}_m}. \label{eq:opt_alpha}
\end{equation}
The former is guaranteed to exist since $H_n(\mathbf{e})$ is positive definite, i.e., $\mathbf{r}_m^\mathrm{T}\mathbf{R}_n\mathbf{E}_n\mathbf{R}_n^\mathrm{T}\mathbf{r}_m > 0$ holds. Projecting $\hat{\alpha}_m$ onto the unit interval by the projection operator $P: (-\infty, \infty) \rightarrow [0, 1]$ gives rise to the solution $\alpha^*_m$ of \eqref{eq:alpha_opt} in closed form
\begin{equation}
    \alpha^*_m = P\left(-\frac{(\mathbf{b}_m - \mathbf{e}_{0,n})^\mathrm{T}\mathbf{R}_n\mathbf{E}_n\mathbf{R}_n^\mathrm{T}\mathbf{r}_m}{\mathbf{r}_m^\mathrm{T}\mathbf{R}_n\mathbf{E}_n\mathbf{R}_n^\mathrm{T}\mathbf{r}_m}\right).
\end{equation}
Since $P$ is not continuously differentiable, we approximate $P$ by
\begin{equation}
    \hat{P}(\alpha) = \alpha \; \Phi(\alpha) - (\alpha - 1) \; \Phi(\alpha - 1)
\end{equation}
where $\Phi$ refers to the smooth approximation of the Heaviside function 
\begin{equation}
    \Phi(\alpha) = \frac{1}{1 + \exp(-c \alpha)}
\end{equation}
and $c \in \mathbb{R}_{>0}$ is a scaling parameter. For $c \to \infty$ the function $\hat{P}$ converges towards $P$. Both, $\hat{P}$ and $P$ are depicted in Figure \ref{fig:alpha} for $c=20$. For instance, for the former parameter choice, the maximum absolute error of  $\alpha^*_m$ amounts to $1.13\cdot 10^{-2}$.

Considering static collision avoidance, formulated in equation \eqref{eq:static_collision_k}, we follow a similar approach as explained before by approximating objects with convex bodies, i.e. by spheres or ellipsoids depending on the geometry of the considered object. In our setup, the table represents a static object, so that there is a risk that the robot $i$ chooses a trajectory bellow or through the table in order to avoid another robot $j$. For this purpose it is sufficient to formulate a plane along the table and restrict the intersection of the basis of each link $\mathbf{b}_m$ of the robot $i$ with the plane by the height of the table denoted as vector $\mathbf{z}=[0 \ 0 \ z_\textrm{T}]$, i.e.
\begin{equation}
    \mathbf{b}_m \geq \mathbf{z} + \mathbf{z}_\textrm{min},
\end{equation}
with an offset $\mathbf{z}_\textrm{min}$. In case of the gripper, which is attached to the end effector, an additional offset equal to the length of the gripper should be considered.   

\subsection{Inter-robot collision avoidance with ELS method for two robots}
In order to ensure collision free trajectory of a robot in a multi-robot setting, it is necessary to encompass the whole geometry of a robot, as described in the previous section \ref{ELS}. In case of 2 manipulators with $N=6$ degrees of freedom, we approximate the robot $i$ for which the DMPC problem is solved by $N_\textrm{L}=8$ line segments, starting from the basis and ending by the gripper. We approximate the neighboured robot by $N_\textrm{E}=5$ ellipsoids, encapsulating the basis, subsequent three links (Shoulder, Elbow, Wrist $2$) and the end-effector including the gripper, which is referred to as Wrist $3$. By ensuring no intersections between any line segment with any ellipsoid, this results in formulating $N_\textrm{dyn}=N_\textrm{L} \cdot N_\textrm{E} = 40$ collision constraints for every time step of the prediction horizon. As the ellipsoids should be chosen large enough to contain at least the diameter of the neighboured link, it is sufficient to omit the $3$ short line segments connecting two joints, i.e. line segment connecting basis and shoulder joints, shoulder and elbow joints as well as wrist $1$ and wrist $2$ joints. This results in formulating $N_\textrm{dyn}=25$ collision constraints for a single time step.

However, the former only serves as an upper bound. Geometrically it is impossible to position robots in a pick \& place setup, where all constraints need to be taken into account, as the manipulators operate in a certain distance to each other. In the work at hand, each manipulator (may it be the two, three or four robot setup) is positioned on top of a flexible module which might be combined arbitrarily with other modules to form larger formations. The minimum distance between the robots dictated by the modules ensures, for example, that one robot cannot touch the base of the other robot. Furthermore, the shoulders of both robots are also not able to collide. Thus, the formulation can be reduced to $N_\textrm{dyn} = 12$ collision avoidance constraints for single time step, which are sufficient for a safe interaction between two robots with $N=6$ degrees of freedom. Those are listed in Table \ref{tab:dyn_constr_2Robots}. For instance, choosing a prediction horizon length of $N_\textrm{p}=20$ results in a total of $240$ constraints to be considered for each manipulator by the DMPC. In case that the robots are placed very close to each other, similar assumptions can be made, where certain constraints can be omitted as well. The former is a matter of the geometric composition of the robots and therefore setup-dependent and might be determined in a pre-processing step.
\begin{table}[h!]
    \centering
        \caption{Intersection of links for formulating collision avoidance constraints in case of two robots $j$ and $i$.}
        \begin{tabular}{ |p{1.4cm}|p{4cm}|p{1.5cm}|}
         \hline
        Robot $j$  & Robot $i$ & $N_\textrm{dyn} \cdot N_\textrm{p}$\\
        (Ellipsoids) & (Line Segments) &  \\
        \hline
        Shoulder &  Wrist $2$, Wrist $3$ & $2 \cdot N_\textrm{p}$\\
        Elbow & Elbow, Wrist $2$, Wrist $3$ & $3 \cdot N_\textrm{p}$ \\
        Wrist $2$  & Elbow, Wrist $2$, Wrist $3$ & $3 \cdot N_\textrm{p}$ \\
        Wrist $3$ & Shoulder, Elbow, Wrist $2$, Wrist $3$ & $4 \cdot N_\textrm{p}$ \\
        \hline 
        \end{tabular}
    \label{tab:dyn_constr_2Robots}
\end{table}

\section{Detecting and resolving deadlocks for robotic manipulators}\label{sec:deadlock}
Deadlocks occurring in a setup of multiple robots is a well-known problem in the field of mobile robots, UAVs and robotic manipulators \cite{jager2001decentralized}, \cite{tallamraju2018decentralized}.  The problem may arise if one or more robotic manipulators block each other, effectively preventing each other from reaching their target state. In our case, the solution of the optimization problem \eqref{eq:DMPC} is at a Nash equilibrium, when a deadlock occurs. Deviating from the optimal solution would increase the cost of an agent and, therefore, would not be an optimal strategy in a non-cooperative game. To this end, resolving deadlocks requires a supervisory instance, i.e. a coordinator, and some sort of information exchange between the robotic manipulators and the coordinator. In addition, reliably and temporally detecting deadlocks between a group of robots that are in deadlock, is a challenging task. Our approach regarding resolving deadlocks, which will be presented in the following, was inspired by the work of Tallamraju et al. \cite{tallamraju2018decentralized}.\\

We propose an approach involving a local deadlock detection, where each robotic manipulator $i$ checks by itself if it is currently in a deadlock and sends the information to the supervisory instance, i.e. the coordinator. The coordinator as previously introduced in Section \ref{control_structure} resolves an occurred deadlock. A manipulator $i$ detects a deadlock if certain conditions are true, i.e. if the change of joint velocities over the prediction horizon is very small, meaning that manipulator $i$ is slowed down and can not move further
\begin{equation} \label{eq:vel_deadlock}
    \left\lVert \dot{\mathbf{q}}^{* \; N_\textrm{p}}_i - \dot{\mathbf{q}}^{* \; 0}_i \right\rVert \leq \varepsilon_v
\end{equation}
and, at the same time, the deviation of the robot's measured state $\mathbf{x}^\textrm{s}_i$ and the desired state $\mathbf{x}^\textrm{f}_i$ is sufficiently large 
\begin{equation} \label{eq:state_deadlock}
    \left\lVert \mathbf{x}^\textrm{s}_i - \mathbf{x}^\textrm{f}_i \right\rVert \geq \delta_x. 
\end{equation}
If both conditions \eqref{eq:vel_deadlock} and \eqref{eq:state_deadlock} are satisfied, a deadlock is detected and deadlock parameter $\gamma_{D,i} \in \{0, 1\}$ is set to $\gamma_{D,i}=1$ and subsequently send to the coordinator. In case of no deadlocks the deadlock parameter takes a value of zero. Besides that, each manipulator $i$ provides the information to the coordinator about its current and target states in each time step. Should a deadlock be detected for any of the $M$ manipulators, the coordinator computes the minimum distance between the links of the robots and determines the smallest distance between the robots. This step belongs to clustering step, where the coordinator determines which robots belong to a group where deadlock has been detected. Then, all manipulators of this group are deactivated except for the manipulator that is closest to its desired state. This procedure ensures, that only manipulators that are in a deadlock are deactivated, whereas all other robots in the workspace are not restricted in their movement. We propose to move the deactivated robots to their neutral position $\mathbf{x}^\mathrm{D}$, which allows the active robot to find a path to its target. Therefore, the coordinator sets the resolving parameter $\gamma_{\mathrm{R},i}=0$ if a robot $i$ should move to its neutral pose. Once the active robot overcomes the deadlock, the deactivated robots are activated again towards their former targets and finish their tasks. \\

In order to classify which robots are currently in a deadlock, a clustering of robots into groups is necessary. We propose therefore the following algorithm, which is described bellow in order to cluster robots into groups that are in deadlock and groups that are not restricted in their movement.
\begin{algorithmic}[1]
		\STATE {Every robot $i$ is placed into an individual cluster $\mathcal{C}_i$}
		\IF {$\gamma_{D,i}=1$}
		\STATE {Check which robots are in the neighbourhood}
		    \FOR {$j=1$ \TO $M$}
		    \IF {$i \neq j$ \& $\textrm{dist}(R_i, R_j) \leq  d_\textrm{min} $}
		    \STATE {Add robot $R_j$ to the cluster $\mathcal{C}_i$}
		    \ELSE 
		    \STATE {Robot $R_j$ remains in its own cluster}
		    \ENDIF
		    \ENDFOR
	    \ENDIF
		 \STATE Check the smallest residuum for all clusters
		 \FORALL{ $R_i \in \mathcal{C}_i$}
		 \IF {$\min_{R_i \in \mathcal{C}_i}{res(R_i)} < \varepsilon_\textrm{res}$}
		 \STATE Robot $R_i$ with smallest residuum remains active
		 \STATE All the other robots receive a $\gamma_{R,i}=0$ and a neutral pose as new target pose $\mathbf{x}_i^\textrm{f}= \mathbf{x}^\textrm{D}$
		 \ELSE
		 \STATE {Reset cluster} 
		 \STATE {All robots are active again, i.e., $\gamma_{R,i}=1$}
		 \ENDIF
		\ENDFOR
	
\end{algorithmic} 

\section{Results} \label{sec:results}
\subsection{Simulation setup and controller parametrization}
The multi-robot setup is built in the robotic simulation environment \highlight{Gazebo} \cite{koenig2004design} with simulated collaborative robotic manipulators UR3 from \highlight{Universal Robots} with $N=6$ degrees of freedom each. \highlight{Gazebo}  provides an interface to control the robots using the Robot Operating System (ROS) \cite{ros}. In this paper we use the distribution \highlight{ROS Noetic}. The communication is established through the ROS action client to the \highlight{Universal Robot} ROS driver. Therefore, a velocity controller hardware interface is applied. The ROS interface allows an easy replacement of the simulation environment in \highlight{Gazebo} by an experimental test bed. 

The control algorithms are implemented in \highlight{Matlab} using \highlight{CasADi} \cite{Andersson2019} for setting up the non-linear program for the DMPCs. The merit of \highlight{CasADi} is its automatic differentiation capability, i.e., \highlight{CasADi} computes the first and second order derivatives of the cost function and constraints using automatic differentiation. We use the interior point solver \highlight{IPOPT} \cite{wachter2006implementation}  to solve the optimization problem and apply \highlight{MA27} \cite{hsl2007collection} to solve the underlying linear system. We set the maximum number of iterations to $1000$ and an acceptable tolerance of $10^{-8}$. In addition, \highlight{CasADi} is instructed to pre-compile the optimization problems using just-in-time compilation. We choose a sampling time of $T_\textrm{s}=200 \ \unit{ms}$.  

The model predictive controllers run in parallel on a computer with an Intel i$7$-$11800$H CPU at $2.30$ GHz using $32$ GB RAM under Ubuntu $20.04$. The coordinator is running on the same computer and communicates the computed optimal trajectories with the model predictive controllers via the UDP protocol. The trajectories between the robots are exchanged via the UDP protocol as well.

The weighting matrices of the DMPCs are chosen as  
\begin{align*}
    \mathbf{Q}_i^\textrm{x}&=\textrm{diag}(1,1,1,0.2,0.2,1,1,1,1,0.1,0.1,0.1),\\
    \mathbf{Q}^\textrm{f}_i&=10 \cdot \mathbf{Q}_i^\textrm{x},\quad 
    \mathbf{R}_i^\textrm{u}=\mathbf{I}^{6 \times 6}  \quad \textrm{and} \quad    \mathbf{R}_i^\textrm{d}=\mathbf{I}^{6 \times 6}. 
\end{align*}
Joint positions and velocities of the three wrist links (connecting the last three joints) are penalized less to allow for a greater freedom of motion. The absolute values of joint velocities of an UR$3$ manipulator are limited to
\begin{equation*}
    [\pi, \pi,\pi,2\pi,2\pi,2\pi] \ \unit{\frac{rad}{s}}.
\end{equation*}
In addition, the absolute values of accelerations are limited to
\begin{equation*}
    [\pi, \pi,\pi,2\pi,2\pi,2\pi] \ \unit{\frac{rad}{s^2}}.
\end{equation*}
The parameters for the deadlock algorithm are set as follows 
\begin{align*}
         \varepsilon_v =1.5 \cdot 10^{-3} \ \unit{\frac{rad}{s}}, \quad \delta_x = 1.2\cdot10^{-2} \ \unit{rad}, \\
         d_\textrm{min} = 0.2 \ \unit{m} \quad \textrm{and} \quad \delta_\textrm{tol} = 4 \cdot10^{-2} \ \unit{rad}. 
\end{align*}

\subsection{Input delays of closed-loop non-linear model predictive control}
The number of active dynamic collision constraints summarized in equation \eqref{eq:dynamic_collision_k} change dynamically, as not all possible collisions can occur at a time step $k$. The number of active collision constraints considerably affects the computation times of the non-convex optimization problem $\eqref{eq:DMPC}$. However, the sampling time $T_\textrm{s}$ cannot be increased arbitrarily, otherwise tunneling will occur resulting in undetected collisions. For this reason, it is necessary to choose a sufficiently small sampling time and to account for the non-negligible computation times as explained by Grüne and Pannek \cite{grune2017nonlinear}.

\subsection{Validation of the motion plan algorithm}
In this section, we demonstrate the flexibility of our approach by arranging the robot modules into different constellations of two, three, and four robots to study the optimal trajectories and computation times for pick and place tasks in more detail. Each manipulator is placed on top of a module of a height $z_\textrm{T} =1.107 \ \unit{m}$. The task for each robot $i$ is to grasp an object in a common workspace and place it into an assigned tray. Each tray is shared by two manipulators. The objects are randomly placed in a common workspace via random sequential adsorption (RSA) by considering additional reachability constraints of the robots. \\
\begin{figure}[h!]
    \centering
    \includegraphics[width=0.49\textwidth]{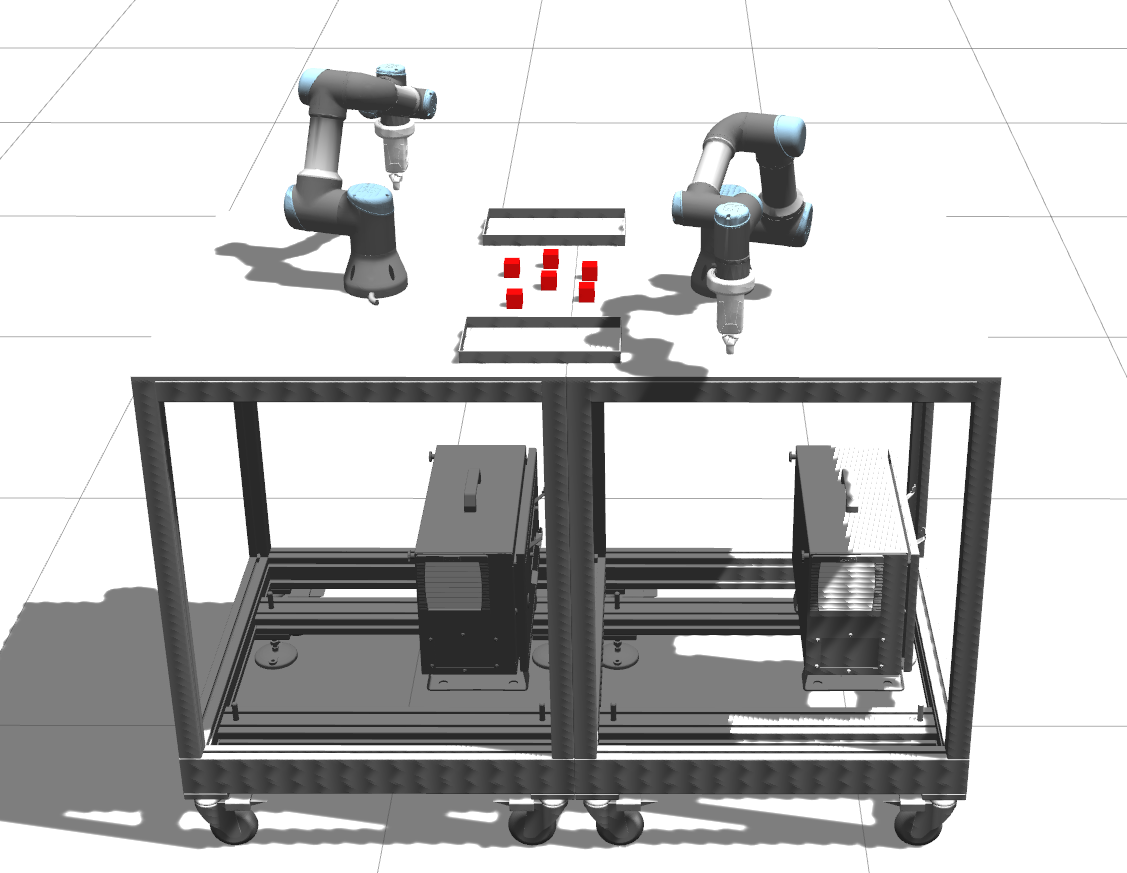}
    \caption{Simulation setup with $2$ modules of UR$3$ robots.}
    \label{fig:setup_2Robots}
\end{figure}

\begin{figure}[h!]
   \centering
    \begin{subfigure}{0.24\textwidth}
        \includegraphics[width=\textwidth]{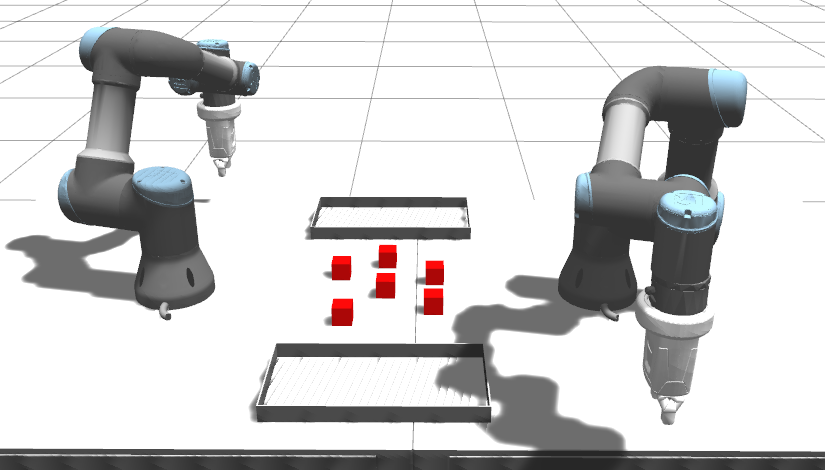}
        \caption{$t= 0 \ \unit{s}$}
    \end{subfigure}
    \begin{subfigure}{0.24\textwidth}
        \centering
        \includegraphics[width=0.8\textwidth]{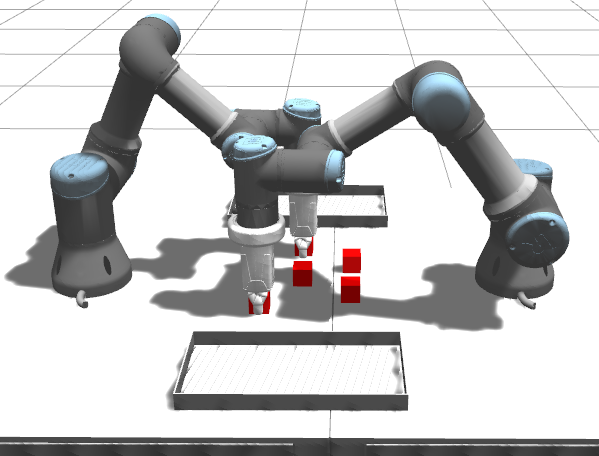}
        \caption{$t= 10 \ \unit{s}$}
    \end{subfigure}
        \begin{subfigure}{0.24\textwidth}
        \includegraphics[width=\textwidth]{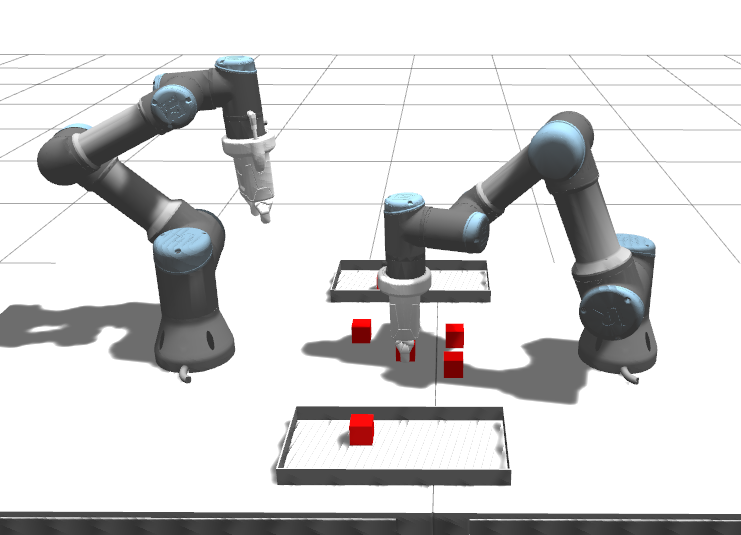}
        \caption{$t= 29 \ \unit{s}$}
    \end{subfigure}
    \begin{subfigure}{0.24\textwidth}
        \centering
        \includegraphics[width=0.8\textwidth]{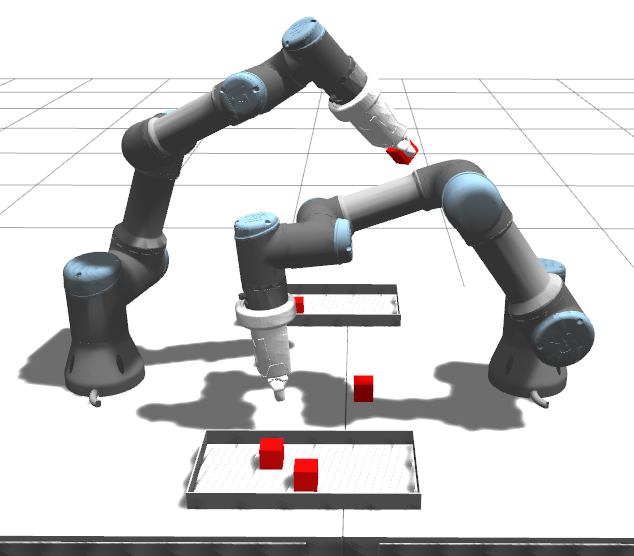}
        \caption{$t= 54 \ \unit{s}$}
    \end{subfigure}
        \begin{subfigure}{0.24\textwidth}
        \centering
        \includegraphics[width=0.9\textwidth]{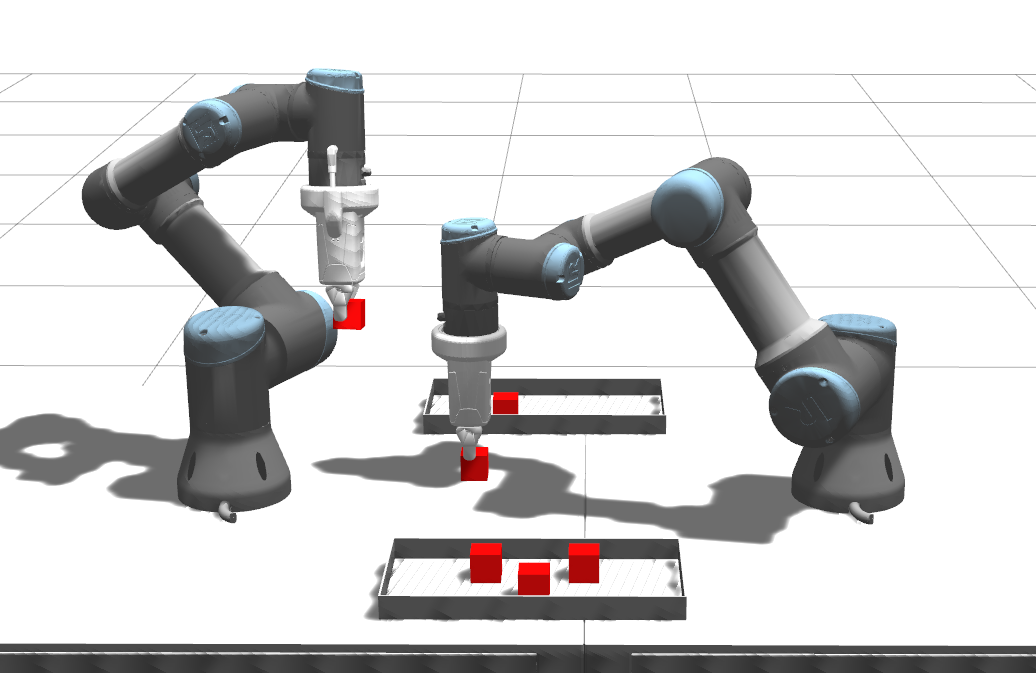}
        \caption{$t= 73 \ \unit{s}$}
    \end{subfigure}
        \begin{subfigure}{0.24\textwidth}
        \includegraphics[width=\textwidth]{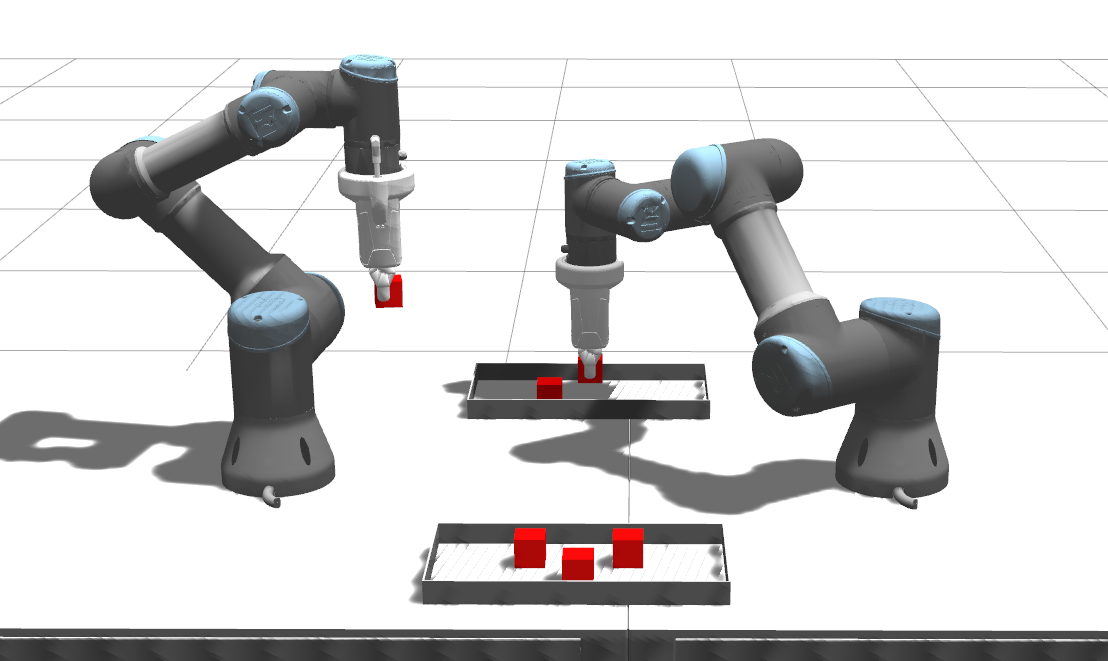}
        \caption{$t= 82 \ \unit{s}$}
    \end{subfigure}
    \begin{subfigure}{0.24\textwidth}
        \centering
        \includegraphics[width=0.8\textwidth]{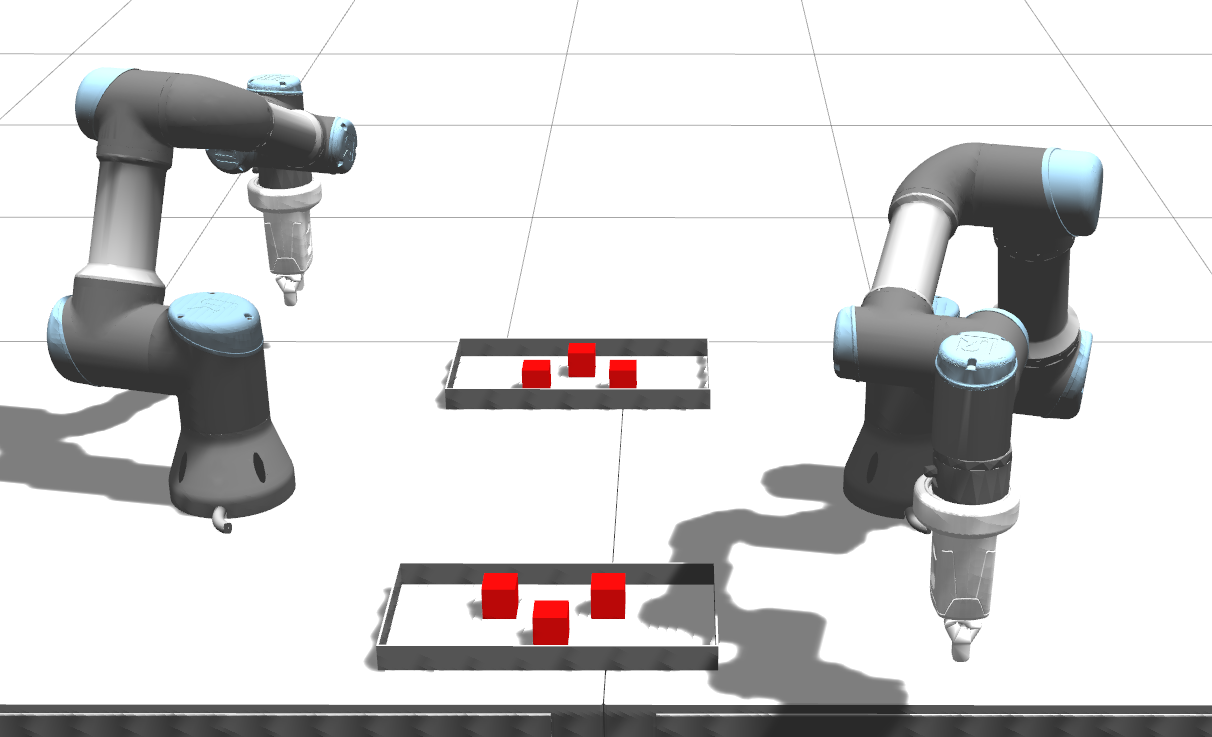}
        \caption{$t= 110 \ \unit{s}$}
    \end{subfigure}
    \caption{Selected time frames from Gazebo simulation for Use Case $1$.}
    \label{fig:Gazebo_2Robots}
\end{figure}

In our first setup, we consider two robots, two trays and six objects, shown in Figure \ref{fig:setup_2Robots}. The randomly distributed objects in the common workspace can be reached by both robots. Similarly, each tray can be served by both robots and contains three slots. The robots are placed close to each other, so that inter-robot collisions are imminent. In first step, the tasks are equally distributed among the two robots by providing a sequence of setpoints to each manipulator. Each robot's task is to place three of the randomly distributed objects into assigned trays. We conduct five use cases with different positions of objects in the shared workspace.  In order to analyze the influence of the prediction horizon length $N_\textrm{p}$ on performance and computation times, we choose three different prediction horizon lengths $N_\textrm{p} \in \{10, 15, 20\}$. In the following, the results for Use Case $1$ is discussed in more detail. For illustration purposes, several time frames are depicted in Figure \ref{fig:Gazebo_2Robots} for Use Case $1$ and a prediction horizon length of $N_\textrm{p}=20$. At time $t=29 \ \unit{s}$ in Figure \ref{fig:Gazebo_2Robots}c, a deadlock has been resolved, where the robot on the left has been deactivated while allowing the robot on the right to grasp its object. At time $t=54 \ \unit{s}$ in Figure \ref{fig:Gazebo_2Robots}d the robot on the left successfully plans an optimal and collision free motion above its neighbour to reach its target. Later, at time $t=73 \ \unit{s}$ in Figure \ref{fig:Gazebo_2Robots}e the robot on the left moves to its neutral pose so that the robot on the right is able to grasp its object. Both robots have to serve the same tray and therefore can not place their objects simultaneously. Therefore, the robot on the left is deactivated once again at time $t= 110 \ \unit{s}$  until robot on the right finishes its task, shown in Figure \ref{fig:Gazebo_2Robots}f. 

\begin{figure}[h!]
   \centering
   \begin{subfigure}{0.49\textwidth}
        \centering
        \includegraphics[height=0.45cm]{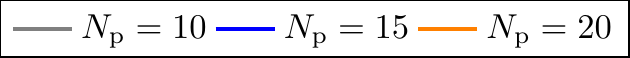}
   \end{subfigure}
    \begin{subfigure}{0.24\textwidth}
        \includegraphics[width=\textwidth]{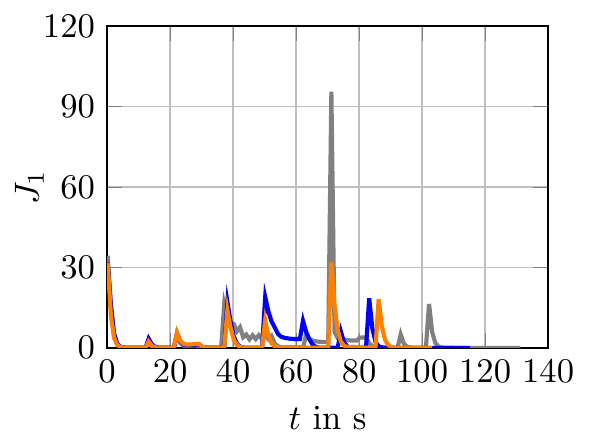}
        \caption{Robot $1$}
    \end{subfigure}
    \begin{subfigure}{0.24\textwidth}
        \centering
        \includegraphics[width=\textwidth]{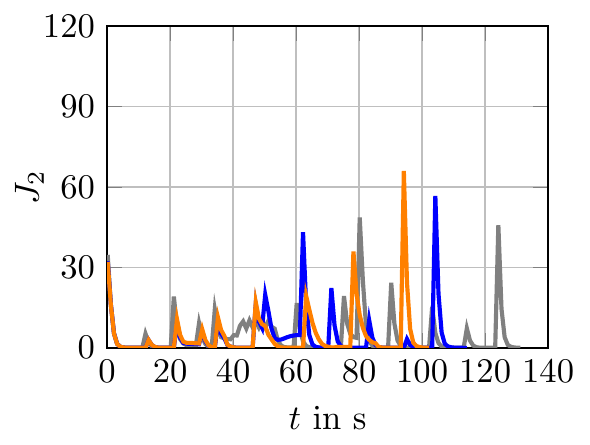}
        \caption{Robot $2$}
    \end{subfigure}
    \caption{Cost function dependence on prediction horizon $N_\mathrm{p}$ for Use Case $1$.}
    \label{fig:costFunc_2Robots}
\end{figure}

The cost functions for both robots are provided in Figure \ref{fig:costFunc_2Robots}. As the cost function punishes the deviation of the current state to the desired state, it rises every time a robot receives a new desired state. 
Furthermore, it can be observed that the execution time, i.e., the time needed to finish all pick and place tasks, reduces with increasing prediction horizon length. For this reason, a prediction horizon length as large as possible is desired that results in faster reactions to upcoming collisions and therefore sooner actions can be taken to avoid them.  

\begin{figure}[h!]
   \centering
   \begin{subfigure}{0.49\textwidth}
        \centering
        \includegraphics[height=0.45cm]{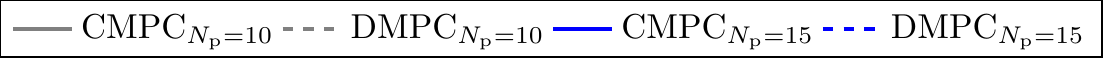}
   \end{subfigure}
    \begin{subfigure}{0.24\textwidth}
        \includegraphics[width=\textwidth]{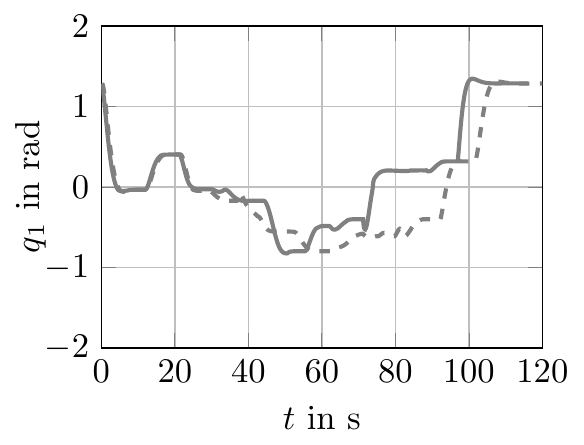}
        \caption{Joint angle $q_1$ of Robot $1$}
    \end{subfigure}
    \begin{subfigure}{0.24\textwidth}
        \centering
        \includegraphics[width=\textwidth]{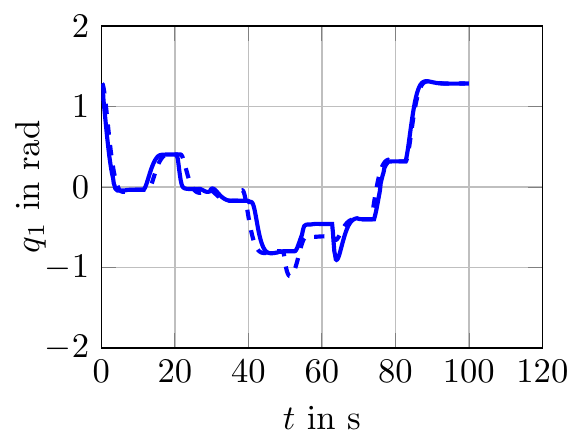}
       \caption{Joint angle $q_1$ of Robot $1$}
    \end{subfigure}
    \begin{subfigure}{0.24\textwidth}
        \includegraphics[width=\textwidth]{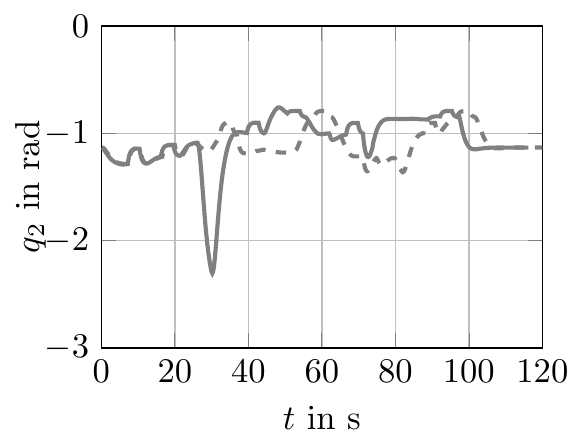}
        \caption{Joint angle $q_2$ of Robot $1$}
    \end{subfigure}
    \begin{subfigure}{0.24\textwidth}
        \centering
        \includegraphics[width=\textwidth]{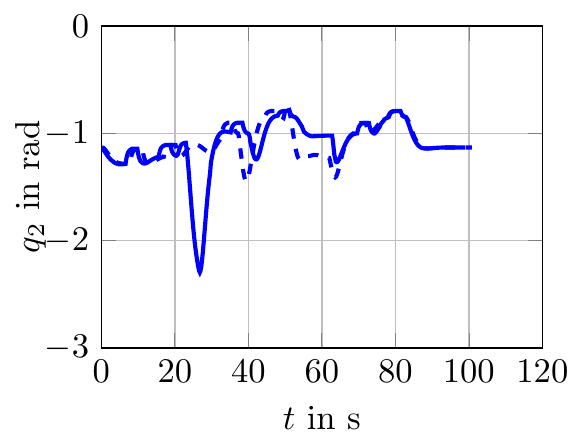}
        \caption{Joint angle $q_2$ of Robot $1$}
    \end{subfigure}
    \caption{Comparing solutions from distributed and centralized MPC for joint angles $q_1$ and $q_2$ with different prediction horizon lengths (Use Case $1$).}
    \label{fig:angles_2Robots}
\end{figure}

Concerning the optimality of the DMPC, we compare the joint angles with the results obtained by centralized MPC (CMPC) for different prediction horizon lengths. In Figure \ref{fig:angles_2Robots} the two joint angles $q_1$ (basis) and $q_2$ (shoulder) are depicted for Use Case $1$, solved in distributed and centralized fashion. For the sake of brevity, the results of the other four joint angles are omitted. Please note that the CMPC for $N_\textrm{p}=20$ is not real time capable as the computation times consistently exceed the sampling time $T_\textrm{s}$. Thus, we restrict to prediction horizon lengths of $N_\textrm{p}=10$ and $N_\textrm{p}=15$. It can be observed, that the difference between the two solutions of CMPC and the DMPC increase with time for $N_\textrm{p}=10$ for both joint angles. For $N_\textrm{p}=15$ the distributed solution follows the solution of the CMPC remarkably well.

\begin{figure}[h!]
   \centering
  \begin{subfigure}{0.49\textwidth}
        \centering
        \includegraphics[height=0.45cm]{graphics/results/2Robots/UseCase1_costFunction_legend.pdf}
   \end{subfigure}
    \begin{subfigure}{0.24\textwidth}
        \includegraphics[width=\textwidth]{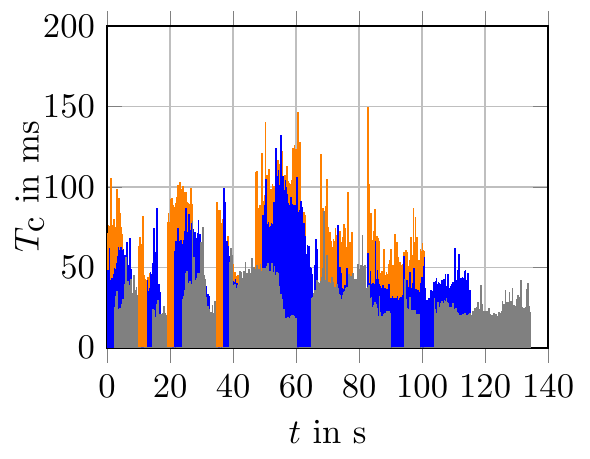}
        \caption{Robot $1$}
    \end{subfigure}
    \begin{subfigure}{0.24\textwidth}
        \centering
        \includegraphics[width=\textwidth]{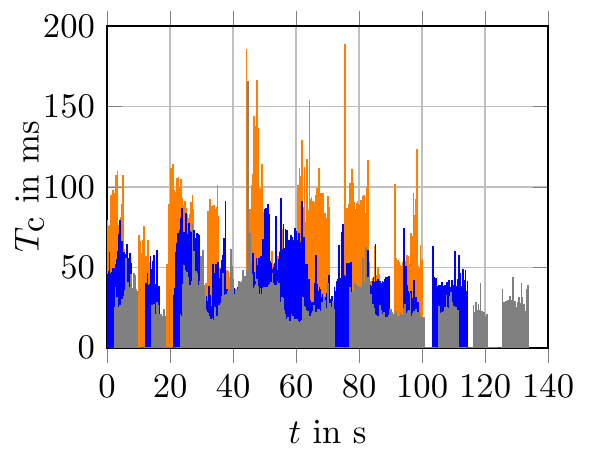}
        \caption{Robot $2$}
    \end{subfigure}
    \caption{Comparison of computation times $T_\mathrm{c}$ to prediction horizon $N_\mathrm{p}$ for Use Case $1$.}
    \label{fig:compTimes_2Robots}
\end{figure}

The computation times for Use Case $1$ are depicted in Figure \ref{fig:compTimes_2Robots}. As expected, the computation time increases with an increasing prediction horizon length $N_\textrm{p}$. Grasping as well as placing procedure are performed without solving the DMPC, due to the fact that once the robot reaches its target it moves down or up within a short time interval. To this end, the gaps in the computation times refer to grasping and placing procedures. To get a better impression of computation times of the DMPC and the CMPC for all use cases, the mean computation times as well as the standard deviations are summarized in Table \ref{tab:compTimes_2Robots}. In case of the DMPC, it can be noticed, that the mean computation times increase superlinearly with an increasing prediction horizon length for all use cases which might be attributed to the direct solver used by \highlight{IPOPT}. Importantly, the standard deviation increases in the same fashion as well. Compared to the computation times obtained by the CMPC, a speed-up factor of more than $2$ has been achieved by solving the problem in a distributed fashion. As mentioned earlier, the CMPC for $N_\textrm{p} = 20$ is not real-time capable and therefore the computation times are omitted.      

\begin{table}[h!]
    \caption{Computation times for $2$ robots.}
    \centering
        \begin{tabular}{ |p{0.4cm}||p{0.7cm}|p{0.3cm}|p{3.3cm} p{1.5cm}|}
         \hline
        Case & $T_s$  & $N_\mathrm{p}$ & \qquad \quad $\textrm{mean}(T_\mathrm{c}) \pm \textrm{std}(T_\mathrm{c})$  & in \unit{ms} \\
        {} & in \ \unit{ms} & {} & \qquad \qquad DMPC & \quad  CPMC \\ 
        {} & {} & {} & Robot $1$ \qquad \qquad  Robot $2$ & Robot 1 \& 2 \\
         \hline
         $1$ & $200$  & $10$    & $31.6 \pm 13.6$ \qquad $28.4 \pm 11.2$ & $74.6 \pm 24.4$  \\
          & {} &  $15$  & $51 \pm 20.3 $ \qquad \ \  $47.7 \pm 15 $ & $116.2 \pm 37.9$ \\
          & {} &20 & $73.9 \pm  23.9$ \qquad $74.2 \pm 24.8$ & \qquad {-} \\
           \hline
         2  & 200 & 10    & $29 \pm  8.3$ \qquad \quad \ $29.2 \pm 10.7 $ & $83.1 \pm 28.7$ \\
          & {} &  15   & $50.9 \pm 17.4 $ \qquad $52.1 \pm  18.4$ & $127.1 \pm 43.1$ \\
          & {} &20 & $72.9 \pm  26.4$ \qquad $73.5 \pm  30.1$  & \qquad {-}  \\
           \hline
            3 & 200  & 10 &  $29.3 \pm  9.7   $ \qquad \  $30.9 \pm 11.3 $ & $77.2 \pm 29.1$\\
          &  {} & 15  & $ 55.5 \pm 23.1  $ \qquad $56.5 \pm 29 $ & $121.1 \pm 49.2$ \\
          & {} &20  & $69  \pm 20.9 $ \qquad \ \ $64.9 \pm 19.3 $ & \qquad {-}  \\
           \hline
            4 & 200  & 10    & $29.1 \pm 10.7 $ \qquad $31.6 \pm 11.5 $ & $71 \pm 25.5$ \\
          &  {} & 15   & $44.4 \pm 19.2 $ \qquad $48.5 \pm 16.4$ & $111.1 \pm 36.2$ \\
          & {} &20 & $72.3 \pm 28 $ \qquad \ \ $67.8 \pm 28.9$ & \qquad {-}  \\
           \hline
            5 & 200  & 10    & $29.2 \pm 9.7 $  \qquad \  $29.3 \pm 8.9$  & $71.1 \pm 21$\\
          & {} &  15  & $49.7 \pm 19.7 $  \qquad $53 \pm 18 $ & $125.5 \pm 36.8$\\
          & {} &20 & $65.4 \pm 22.5 $ \qquad $72.9 \pm 21 $ & \qquad {-} \\
         \hline
        \end{tabular}
    \label{tab:compTimes_2Robots}
\end{table}

\subsection{Benchmark problems with OMPL planners and CHOMP planner}
\begin{figure}[h!]
    \centering
    \includegraphics[width=0.49\textwidth]{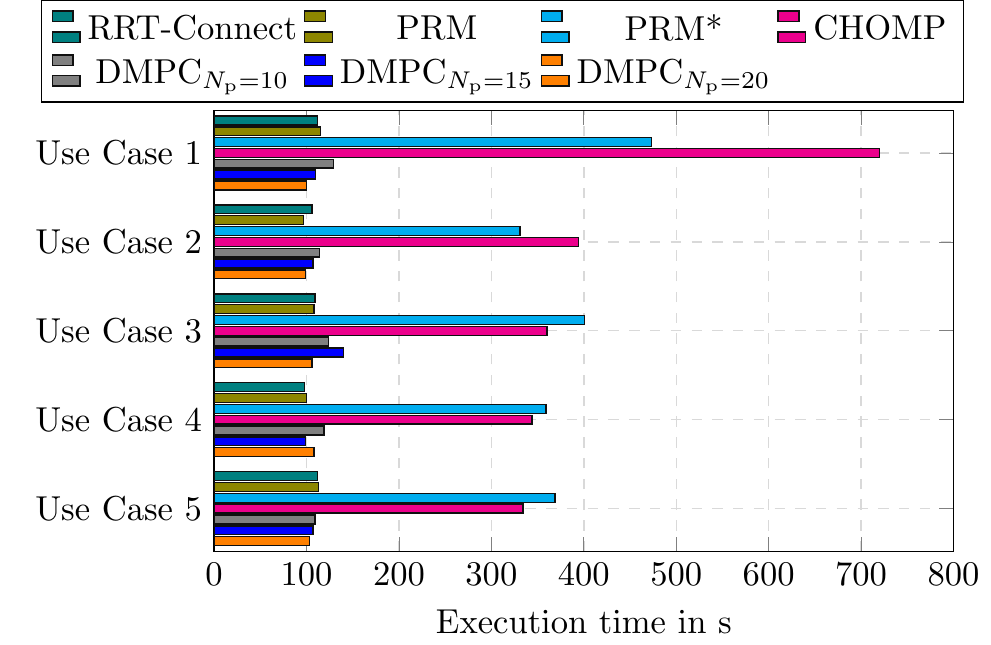}
    \caption{Benchmark with four motion planning methods.}
    \label{fig:errors_bilinear}
\end{figure}

We compare our approach with several sampling-based methods integrated in OMPL, such as RRT-Connect, PRM, PRM$^*$ and the optimization-based method CHOMP with regard to execution times and collision free trajectories for a setup with $2$ robots. The sampling-based methods do not guarantee completeness, i.e. a solution might exist but a planer fails to find one. In addition, planners such as PRM$^*$ need certain amount of time to plan a trajectory. Not restricting the planning time can cause an infinite time to find a solution. Figure \ref{fig:errors_bilinear} represents the execution times, i.e., the times for finishing the pick and place task, for the previously mentioned approaches together with the DMPC for prediction horizons $N_\textrm{p} \in \{10,15,20\}$. The benchmark methods plan the trajectories for two robots at the same time and sends them to the robots executing the trajectories simultaneously. In order to ensure comparability between the benchmark methods and the DMPC, we integrate the same deadlock resolution procedure as for the DMPC, described in Section \ref{sec:deadlock}. In other words, once a planner fails to find a solution for both robots, e.g., simultaneous picking or placing is not feasible due to otherwise occurring collisions (geometrically infeasible poses), one robot is sent to its neutral pose, so that the other one can grasp or place an object. Furthermore, PRM$^*$ needs at least $10 \ \unit{s}$ of planning time to find a solution for every target pose. For this reason, the execution time for the planner considerably exceeds the planning time of PRM and RRT-Connect. We encountered several collisions with the CHOMP algorithm between the robots resulting in a poor performance for cooperation tasks. Especially for Use Case $1$, the planner often failed to find a path. Other algorithms provided jerky and unnecessary motions but still collision free paths for the two robots. To illustrate this effect, we visualized the joint angle $q_1$ of the left robot for Use Case $1$ and $2$, respectively. Please note, that we restrict to the RRT-Connect and PRM planners as PRM* and CHOMP far exceeded the execution time of about $120 \ \unit{s}$. For Use Case $1$, we see an initial good agreement of the planned trajectories with the DMPC for all three examined prediction horizons. From time around $40 \ \unit{s}$ onwards, deviations arise as the robot reaches its goal a bit faster for $N_\textrm{p}=15$ and $N_\textrm{p}=20$. For Use Case $2$, on the other hand, noticeable deviations can be observed for the whole execution time for the two planned trajectories of RRT-Connect and PRM, which correspond to the previously mentioned jerky and unnecessary motions. The former can be attributed to the two heuristic planners which do not guarantee an optimal (in the sense of shortest path or minimal energy) solution.

From the obtained results, the DMPC is in the same range of execution times as RRT-Connect and PRM for all prediction horizons. In addition, our approach is capable of reacting on dynamically chaning environments. In other words, once the one robot's target pose changes the other robot's target is not influenced and it can still re-plan its own motion at any time. This is not the case with the planners studied here.

\begin{figure}[h!]
    \centering
    \begin{subfigure}{0.49\textwidth}
        \centering
        \includegraphics[height=0.45cm]{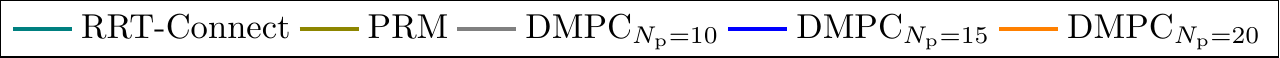}
    \end{subfigure}
   \centering
        \begin{subfigure}{0.24\textwidth}
        \includegraphics[width=\textwidth]{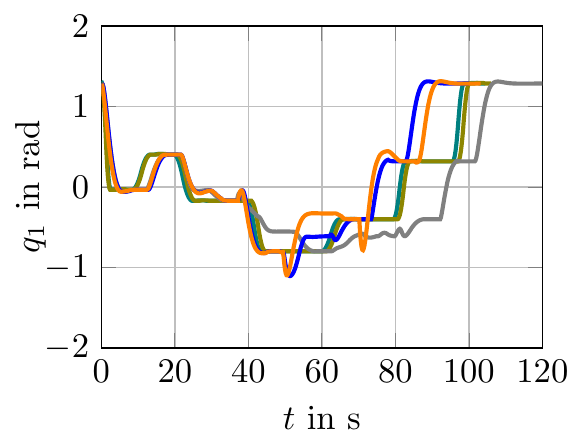}
        \caption{Use case 1}
    \end{subfigure}
    \begin{subfigure}{0.24\textwidth}
        \centering
        \includegraphics[width=\textwidth]{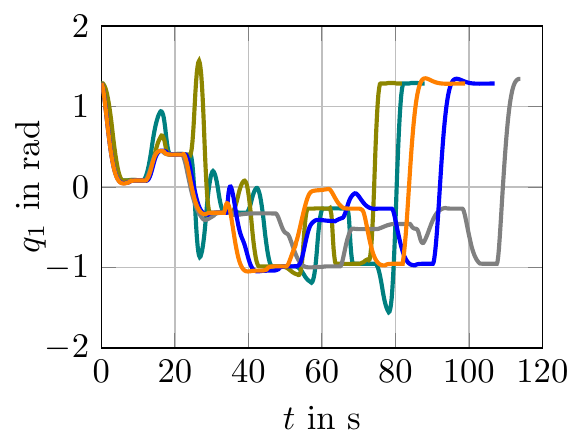}
        \caption{Use case 2}
    \end{subfigure}
    \caption{Comparison of joint angle $q_1$ of the left robot for use cases $1$ and $2$}
    \label{fig:2Rbots_benchmark_Planer_MPC}
\end{figure}

\begin{figure}
    \centering
    \includegraphics[width=0.49\textwidth]{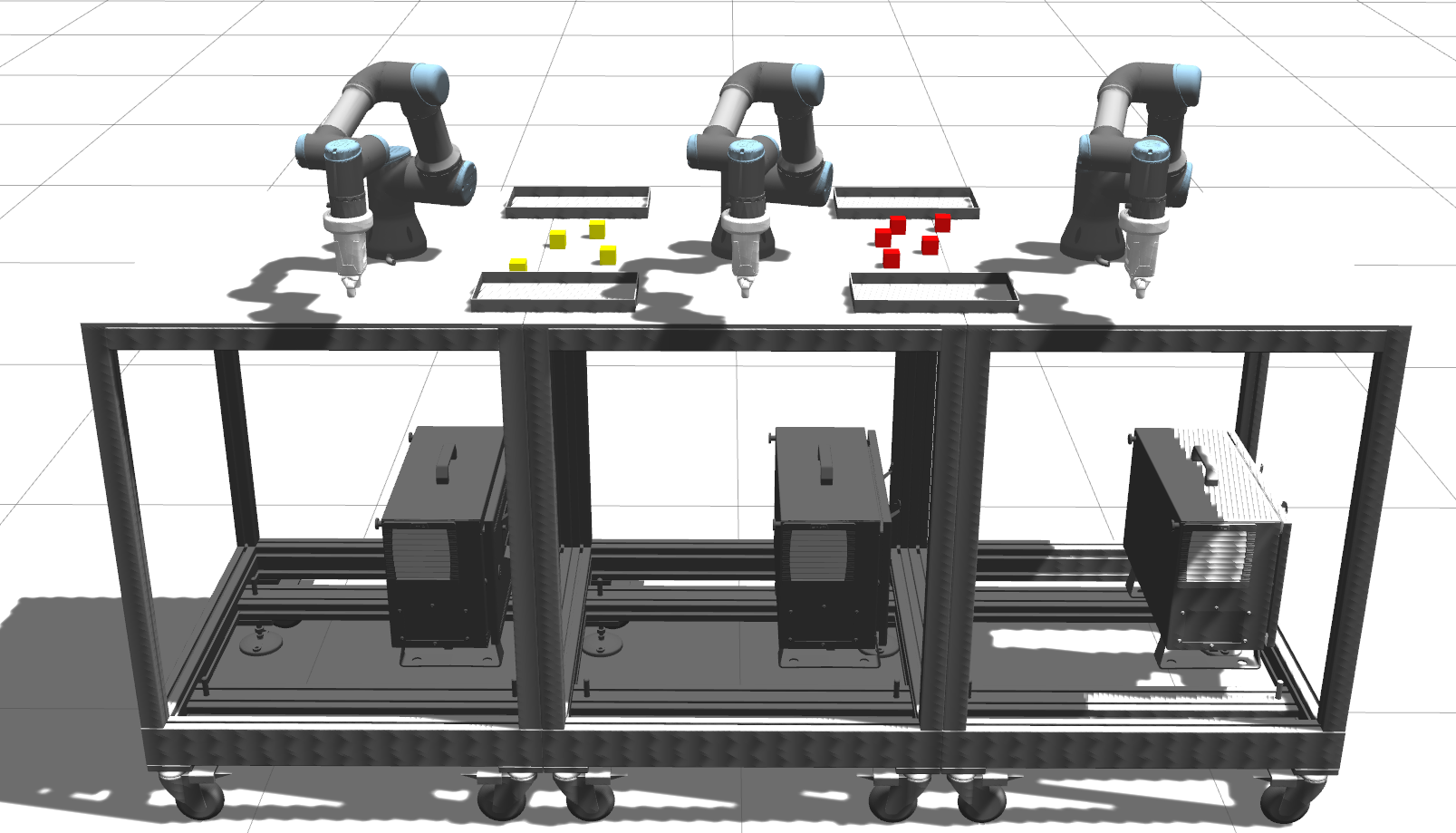}
    \caption{Simulation setup with $3$ modules of UR$3$ robots}
    \label{fig:3Robots_module}
\end{figure}

\subsection{Scalability of the DMPC approach}
\begin{figure}[h!]
   \centering
        \begin{subfigure}{0.24\textwidth}
        \includegraphics[width=\textwidth]{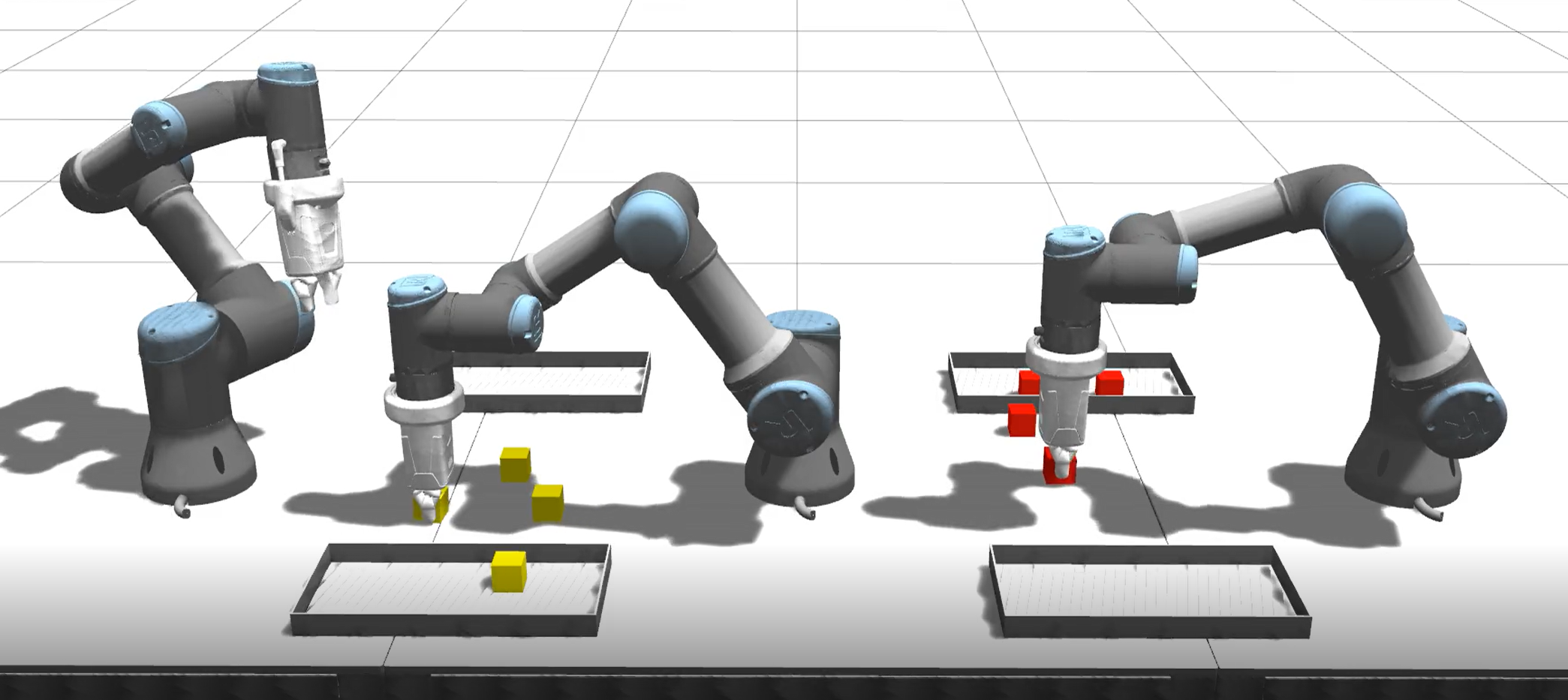}
        \caption{$t= 33 \ \unit{s}$}
    \end{subfigure}
    \begin{subfigure}{0.24\textwidth}
        \centering
        \includegraphics[width=0.9\textwidth]{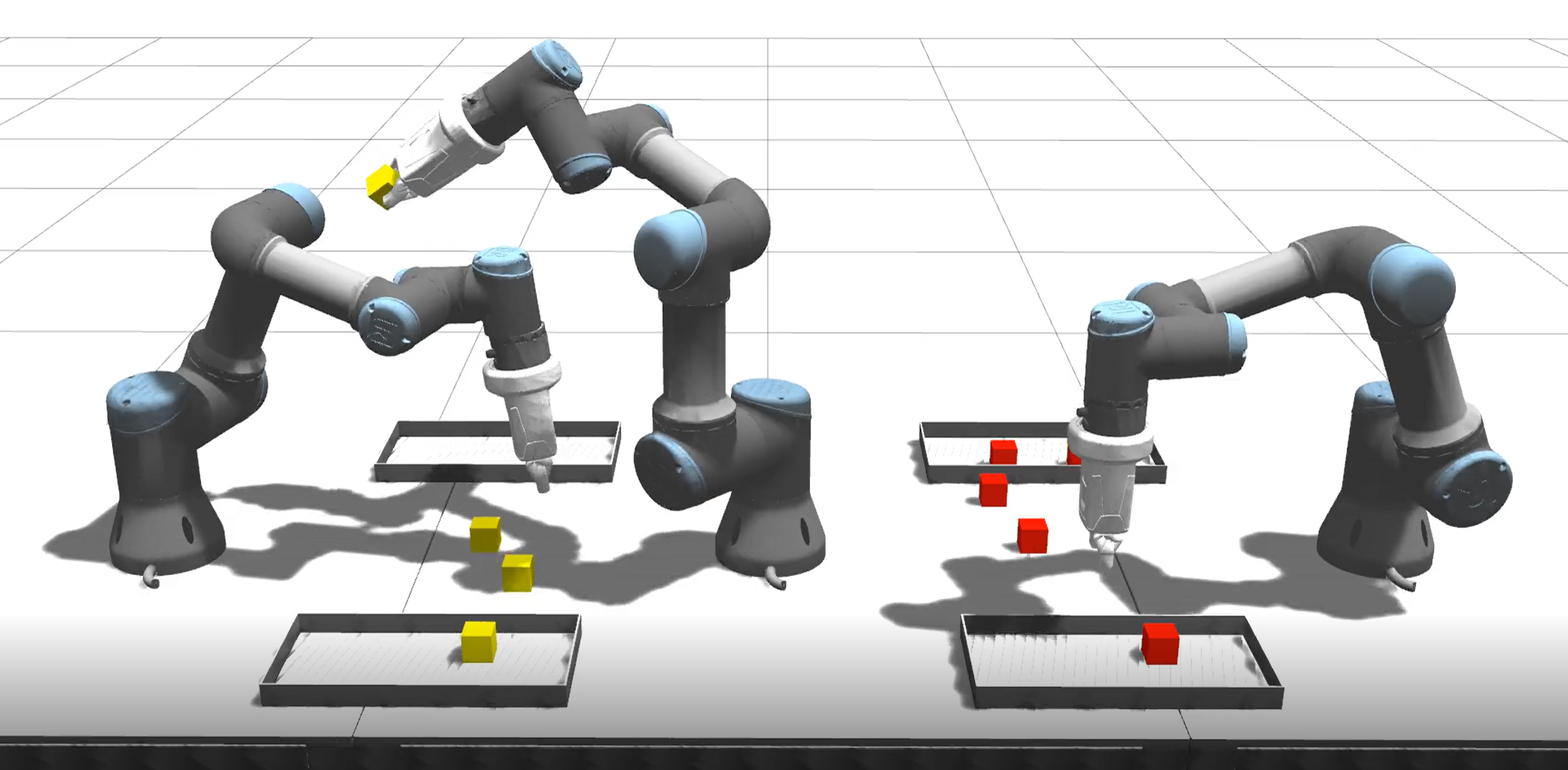}
        \caption{$t= 42 \ \unit{s}$}
    \end{subfigure}
    \caption{Selected time frames from Gazebo simulation for $3$ robots.}
    \label{fig:gazebo_3Robots}
\end{figure}

Subsequently, we study how our approach scales to more than two robot modules. We study setups of three and four robot modules to investigate how the computation times scale with an increasing number of robots. The setup with three modules, each comprising a single robot, are set up in a row, where the outer robots cooperate with the robot in the middle, illustrated in Figure \ref{fig:3Robots_module}. The robot in the middle is performing pick and place tasks in two different workspaces. As before, we consider five individual use cases with randomly placed objects. Each robot is assigned three of these objects which need to be placed in one of the four trays. For all $5$ use cases the robots were able to reliably detect deadlocks and avoid collisions between each other. For sake of brevity, two time steps from Use Case $1$ are illustrated in Figure \ref{fig:gazebo_3Robots}, where deadlock and collision avoidance occurred. The pick and place tasks were successfully performed by the three robots for all use cases.

\begin{figure}[h!]
    \centering
    \includegraphics[width=0.49\textwidth]{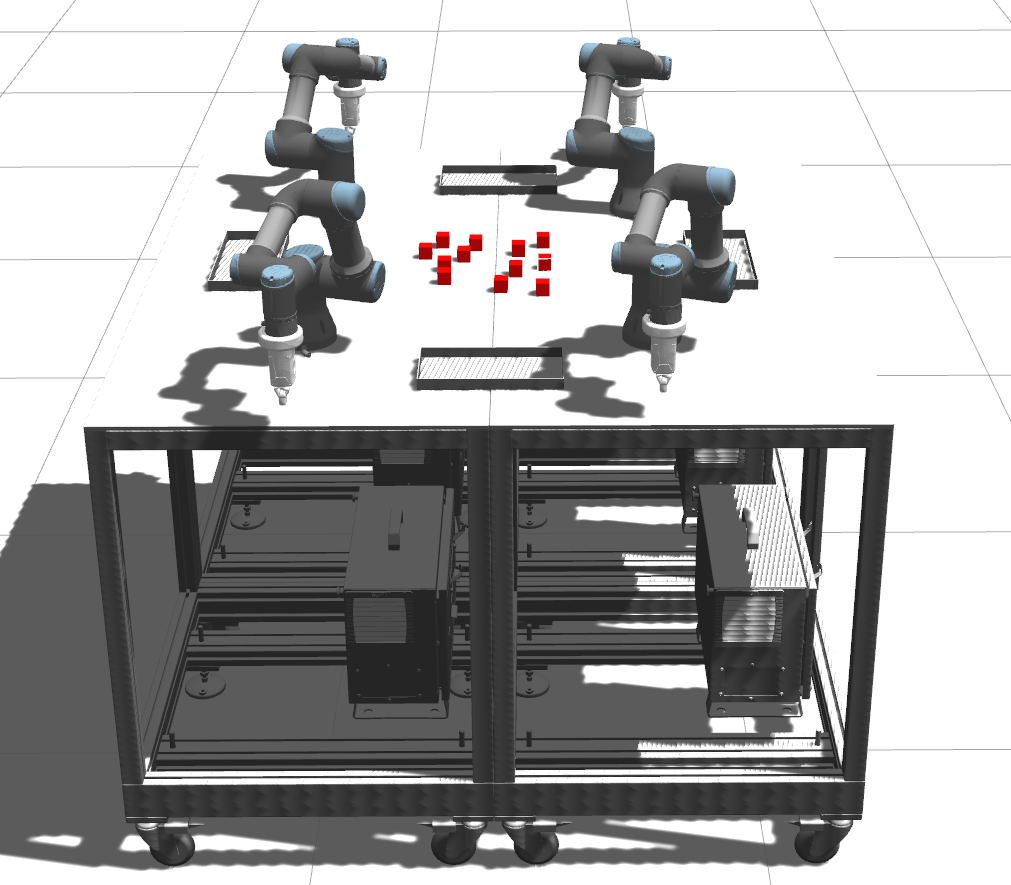}
    \caption{Simulation setup with $4$ modules of UR$3$ robots}
    \label{fig:4Robots_module}
\end{figure}

\begin{figure}[h!]
   \centering
        \begin{subfigure}{0.24\textwidth}
        \includegraphics[width=\textwidth]{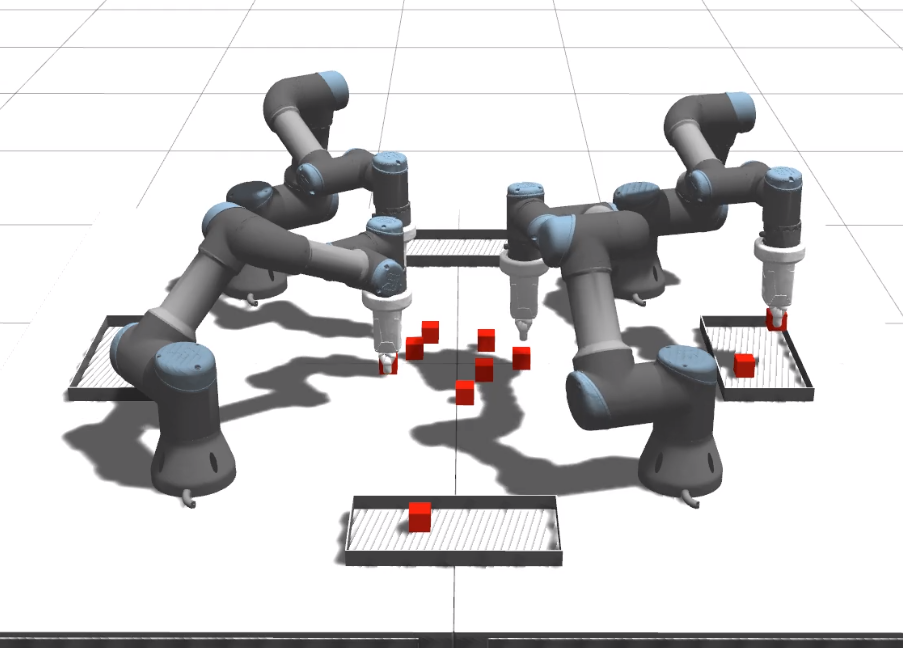}
        \caption{$t= 27 \ \unit{s}$}
    \end{subfigure}
    \begin{subfigure}{0.24\textwidth}
        \centering
        \includegraphics[width=0.9\textwidth]{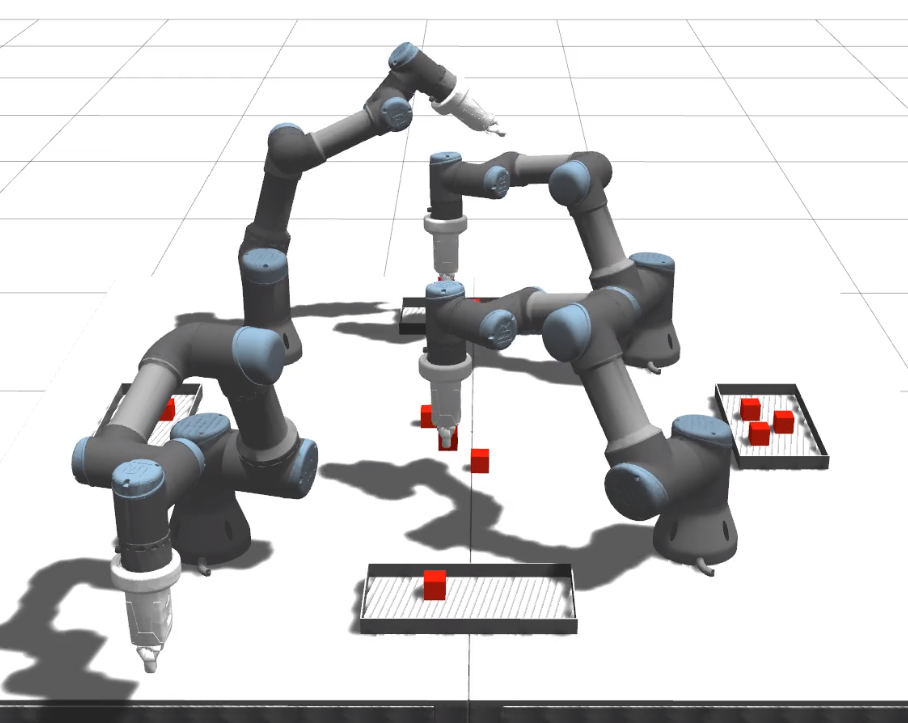}
        \caption{$t= 46 \ \unit{s}$}
    \end{subfigure}
    \caption{Selected time frames from Gazebo simulation for $4$ robots.}
    \label{fig:gazebo_4Robots}
\end{figure}

As the last setup, we consider four modules with four manipulators and $12$ objects in the shared workspace, depicted in Figure \ref{fig:4Robots_module}. In this case, not all objects are reachable by all robots and each tray can only be served by two robots. As before, we study 5 independent use cases with randomly placed object. The robots performed the pick and place task for $12$ objects into the four provided trays while reliably preventing collisions and resolving occurring deadlocks. In Figure ~\ref{fig:gazebo_4Robots}, two distinct time steps are illustrated showing the robots performing the pick and place tasks.

\begin{figure}[h!]
    \centering
    \includegraphics[width=0.375\textwidth]{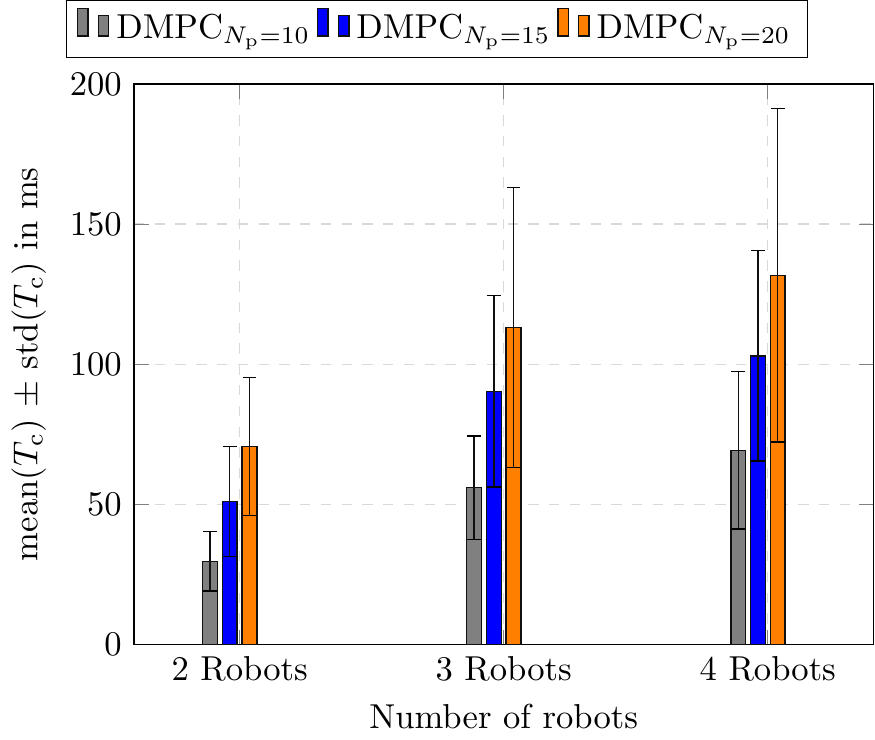}
    \caption{Scaling of computation times with the number of robots.}
    \label{fig:comparison_computation_times}
\end{figure}

Last but not least, we compare mean computation times for the setups of two, three and four robots and its dependency on the prediction horizon length. From Figure \ref{fig:comparison_computation_times} it can be seen, that computation times rise with increasing number of robots. Furthermore, mean computation times increase superlinearly with the prediction horizon lengths. The mean computation times for prediction horizon lengths $N_\textrm{p} \in \{10, 15\}$ do not exceed $100 \ \unit{ms}$ for all number of robots. For $N_\textrm{p} = 20$ the upper boundary of the standard deviation for $3$ and $4$ robots reach around $170$ \unit{ms} and $190$ \unit{ms}, respectively. This limits further increasing the number of robots while fixing the sampling time of $T_\textrm{s}= 200$ \unit{ms} and a prediction horizon length of $N_\textrm{p} = 20$. As previously shown for a setup of $2$ robots, prediction horizon length of $N_\textrm{p} = 15$ is sufficient enough as the solution converges towards the solution of the CMPC. Apart from that, it might be possible to solve optimal trajectories for more than $4$ robots for shorter prediction horizon lengths.  

\section{Conclusion} \label{sec:conclusion}
In this work we introduced a novel motion control algorithm for multiple robotic manipulators. The manipulators were given in terms of multiple modules which were combined into larger structures of 2, 3 and 4 robots. Each robot plans its own collision free trajectory accounting for static and dynamic obstacles. The motion control was realized as a distributed model predictive control (DMPC), formulated as a non-cooperative game. The framework requires a communication between the manipulators for safe interaction with each other. We proposed a novel approach to formulate the collision avoidance constraints. Each robot was approximated by line segments, while the other surrounding robots were approximated by ellipsoids. This formulation allows for a computation of the optimal trajectories in real-time. In a setup of multiple robotic manipulators deadlocks may occur, which is a well known problem in robotics. We proposed an approach, where each manipulator detects if its currently in a deadlock. Based on this information, the introduced coordinator resolves occurring deadlocks without interrupting the motion of not affected robots.\\

The motion control algorithm was validated on different constellations of robotic modules for two, three and four manipulators performing pick and place tasks. The setup was built in the simulation environment \highlight{Gazebo} and controlled by ROS. 
We observed that for cases where robots have to serve the same tray or have to pick objects very close to each other, deadlocks occurred. However, in each case, the manipulators reliably detected deadlocks and the coordinator successfully resolved them. Concerning the optimality of our approach, we compared trajectories of our approach with the centralized solution. The results showed, that with longer prediction horizon, the difference between the solutions decreases and the distributed solution converges towards the centralized one. Finally, a comparison of computation times with centralized MPC as a benchmark showed, that a considerable speed-up is achieved by solving the problem in a distributed manner. Last but not least, we compared our approach with the well-known sampling-based (RRT-connect, PRM, PRM*) and optimization-based planners (CHOMP). Our approach showed a considerable efficiency in both computation times and smoothness of planned trajectories. Finally, we showed by comparing computation times obtained for different numbers of robots, that our framework scales to multiple robotic manipulators. Our approach allows a flexible, real-time capable motion control and trajectory planning for $4$ manipulators with $6$ degrees of freedom and a $200 \ \unit{ms}$ time window. \\
  
In the future, we plan to realize the proposed approach on an experimental testbed for at least two robotic manipulators performing assembly and disassembly tasks. To further increase the efficiency of our approach, it might be beneficial to prevent deadlocks in advance by an intelligent scheduling of the tasks. Such a scheduling algorithm could be realized as a top layer of the proposed control algorithm.

\section*{Acknowledgment}

The authors would like to thank the Ministry of Economics, Transport, Agriculture and Viticulture of the State of Rhineland-Palatinate for financial support within the project "Building a collaborative and cooperative robotics platform-KoKoBot". We thank the anonymous reviewers for their helpful comments.




\bibliographystyle{IEEEtran}
%

\bibliography{Transactions-Bibliography/lit}
%
\vspace{-1.5cm}
\begin{IEEEbiography}[{\includegraphics[width=1in,height=1.2in,clip,keepaspectratio]{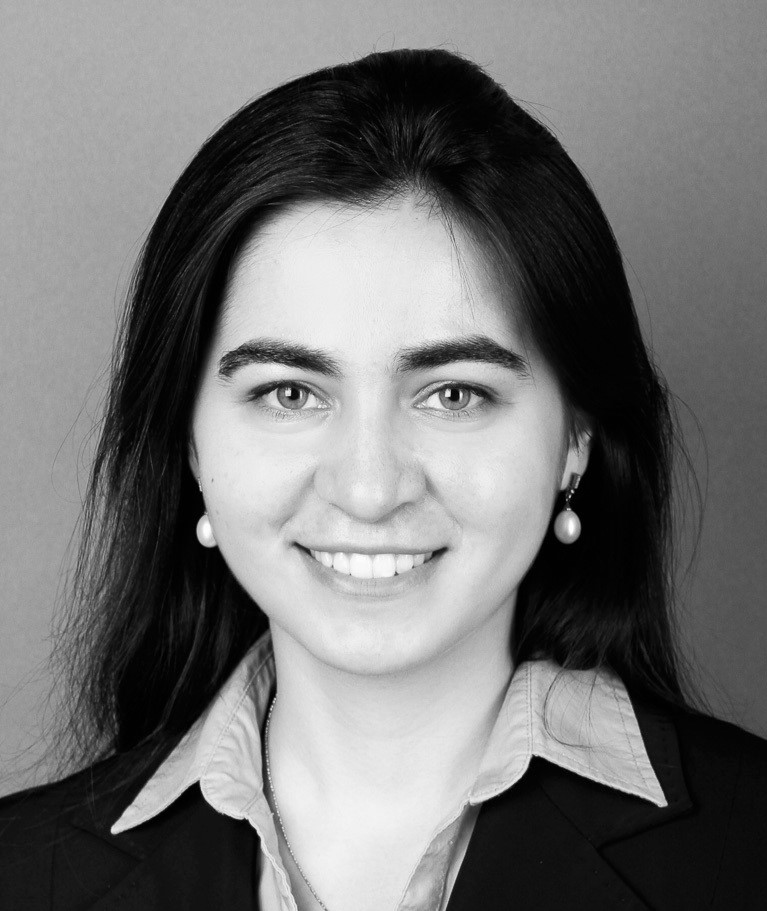}}]{Nigora Gafur}
received the M.Sc. degree in mechanical engineering from Karlsruhe Institute of Technology (KIT), Karlsruhe, Germany, in 2017. Since 2018, she has been working towards the Ph.D. degree in mechanical engineering at Department of Mechanical and Process Engineering at Technische Universität Kaiserslautern (TUK). \\
Her current research interests include dynamics, motion control in robotics and model predictive control. 
\end{IEEEbiography}
\vspace{-1.5cm}
\begin{IEEEbiography}[{\includegraphics[width=1in,height=1.15in,clip,keepaspectratio]{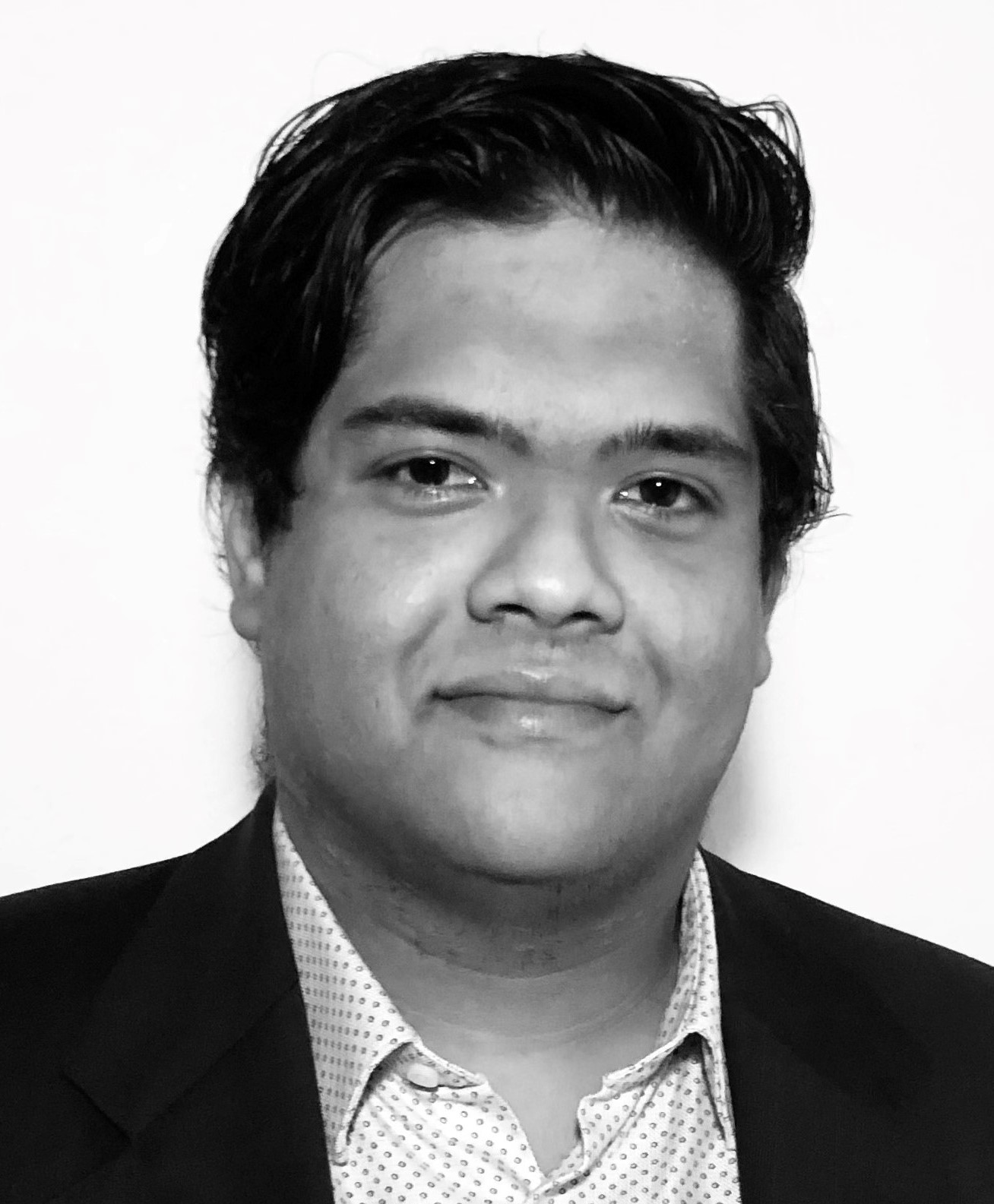}}]{Gajanan~Kanagalingam}
is a graduate student and student research assistant at the Chair of Machine Tools and Control Systems, Department of Mechanical and Process Engineering at Technische Universität Kaiserslautern (TUK). 
\end{IEEEbiography}
\vspace{-1.5cm}
\begin{IEEEbiography}[{\includegraphics[width=1in,height=1.25in,clip,keepaspectratio]{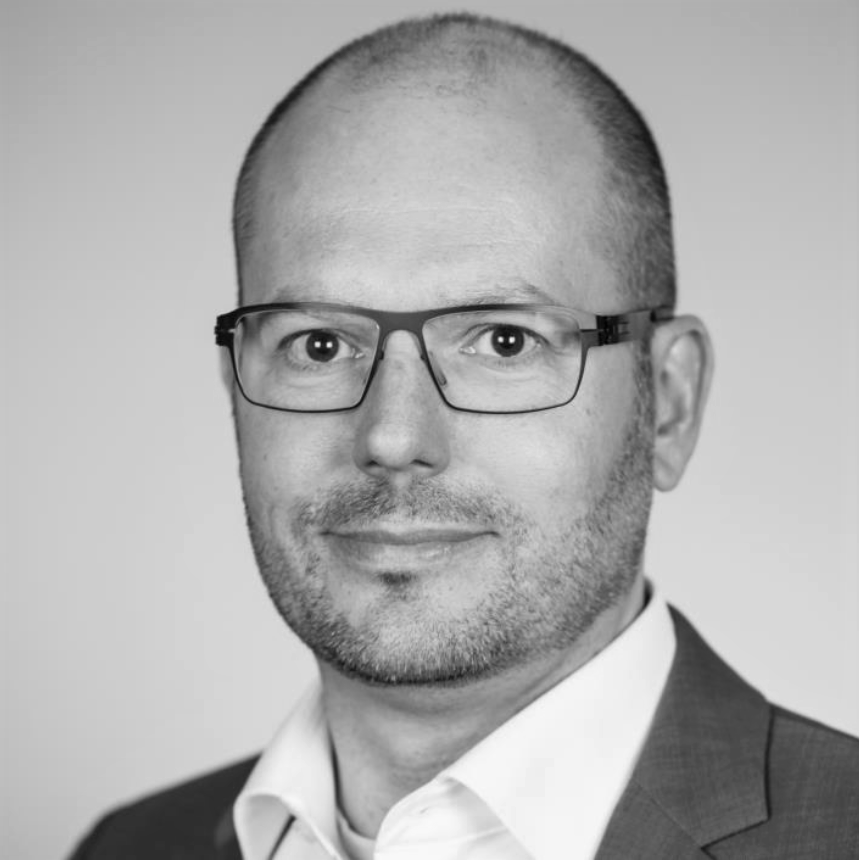}}]{Martin Ruskowski}
received the Diploma degree in electrical engineering from Leibniz University Hannover, Hannover, Germany, in 1996 and the Dr.-Ing. degree in mechanical engineering from Leibniz University Hannover, in 2004. \\
Prof. Dr.-Ing. Martin Ruskowski is Head of the Innovative Factory Systems research department at the German Research Center for Artificial Intelligence (DFKI) and is Chair of the Department of Machine Tools and Control Systems at Technische Universität Kaiserslautern (TUK) and chairman of the board of the technology initiative SmartFactory KL since 2017. His major research focus lies in the development of innovative control concepts for automation, artificial intelligence in automation technology and industrial robots as machine tools. Prior to these positions, Ruskowski held several management positions at industrial firms, most recently as Vice President for Global Research and Development at KUKA Industries Group. 
\end{IEEEbiography}




\end{document}